\documentclass[journal]{IEEEtran}

\usepackage[hidelinks,bookmarks=false,pdfpagelabels=false]{hyperref}
\PassOptionsToPackage{table}{xcolor}
\usepackage{graphics} 
\usepackage{graphicx}
\usepackage{glossaries}
\usepackage{amsfonts}
\usepackage{amsmath}
\usepackage{caption}
\usepackage{algorithm}
\usepackage{algpseudocode}
\usepackage{booktabs} 
\usepackage[table]{xcolor}
\usepackage{siunitx}
\usepackage{soul}
\usepackage{array}
\usepackage{multirow}
\usepackage[short]{optidef}
\usepackage{stackengine}
\ifdefined\pdfsuppresswarningpagegroup
\fi
\usepackage{subcaption}
\usepackage{placeins}

\usepackage{amsthm}
\usepackage{amssymb} 

\newenvironment{compactenum}{%
\begin{list}{\arabic{enumi}.}{%
\usecounter{enumi}%
\setlength{\leftmargin}{0pt}%
\setlength{\itemindent}{1.35em}%
\setlength{\labelwidth}{1.0em}%
\setlength{\labelsep}{0.35em}%
\setlength{\itemsep}{0.2em}%
\setlength{\parsep}{0pt}%
\setlength{\topsep}{0.2em}%
\setlength{\partopsep}{0pt}}}{\end{list}}

\newacronym{mpc}{MPC}{Model Predictive Control}
\newacronym{nmpc}{NMPC}{Nonlinear Model Predictive Control}
\newacronym{rl}{RL}{Reinforcement Learning}
\newacronym{ppo}{PPO}{Proximal Policy Optimization}
\newacronym{sgrl}{SG-RL}{Solver-Gradient Guided Reinforcement Learning}
\newacronym{sgsca}{SG-SCA}{Adaptive PPO Update Scaling}
\newacronym{sglos}{SG-LOS}{Performance-Gated Loss}
\newacronym{sgadv}{SG-ADV}{Advantage Shaping}
\newacronym{sgcrt}{SG-CRT}{Augmented Critic}
\newacronym{gbpl}{GB-PL}{Gradient-Based Policy Learning}
\newacronym{bo}{BO}{Bayesian Optimization}
\newacronym{kkt}{KKT}{Karush-Kuhn-Tucker}
\newacronym{ift}{IFT}{Implicit Function Theorem}
\newacronym{ocp}{OCP}{Optimal Control Problem}
\newacronym{hempc}{HE-MPC}{Human Expert manually parametrized MPC}
\newacronym{bompc}{MOBO-MPC}{Multi-Objective Bayesian Optimization MPC}
\newacronym{rlmpc}{RL-WMPC}{Deep Reinforcement Learning Weights-varying MPC}
\newacronym{diffmpc}{Diff-MPC}{Differentiable MPC}
\newacronym{diffwmpc}{Diff-WMPC}{Differentiable Weights-varying MPC}
\newacronym{mobo}{MOBO}{Multi-Objective BO}
\newacronym{gp}{GP}{Gaussian Process}
\newacronym{pomdp}{POMDP}{Partially Observable Markov Decision Process}

\DeclareMathOperator*{\argmin}{arg\,min}

\title{\LARGE \bf
Accelerating Reinforcement Learning via MPC Solver-Gradient Guidance for Weights-varying MPC
}

\author{Baha Zarrouki, Arslan Thobani, Jasper Hoffmann, Mattia Piccinini,
Rudolf Reiter, Felix Jahncke,\\
Sébastien Gros, Davide Scaramuzza, and Johannes Betz}
\begin{document}
\bstctlcite{IEEEexample:BSTcontrol}

\maketitle
\thispagestyle{empty}
\pagestyle{empty}

\begin{abstract}
In Model Predictive Control (MPC), cost-function weights shape closed-loop behavior, yet changing conditions often make fixed parametrizations suboptimal and motivate context-dependent online adaptation. Learning such policies is difficult because behavior depends implicitly on numerical MPC solutions, producing nonlinear, potentially nonsmooth, long-horizon dependencies on policy parameters. This creates a bias--variance tradeoff: Reinforcement Learning (RL) optimizes realized closed-loop return from environment samples but is sample-inefficient, whereas Gradient-Based Policy Learning (GB-PL) uses low-variance solver gradients from differentiable MPC to optimize surrogate losses on predicted trajectories but can be biased under model mismatch.
%
We propose \textbf{Solver-Gradient Guided Reinforcement Learning (SG-RL)}, a solver-sensitivity augmentation for RL-based online MPC cost-weight adaptation. SG-RL keeps sampled closed-loop return as the objective and uses bounded solver-derived gradients as auxiliary guidance to improve stability and sample efficiency.
We instantiate SG-RL in Proximal Policy Optimization (PPO) with four modular algorithms that inject solver-gradient guidance into actor-update scaling, policy loss, advantage estimation, and value-function learning. On two full-scale autonomous racing platforms with intentional model mismatch, SG-RL reaches PPO's best closed-loop return with up to 70.6\% fewer samples, outperforms GB-PL baselines by at least 54\% in closed-loop return, and generalizes zero-shot to unseen environments.

\end{abstract}


\section{INTRODUCTION}
\label{sec:introduction}
\begin{figure}[!t]
    \centering
    \includegraphics[width=\columnwidth]{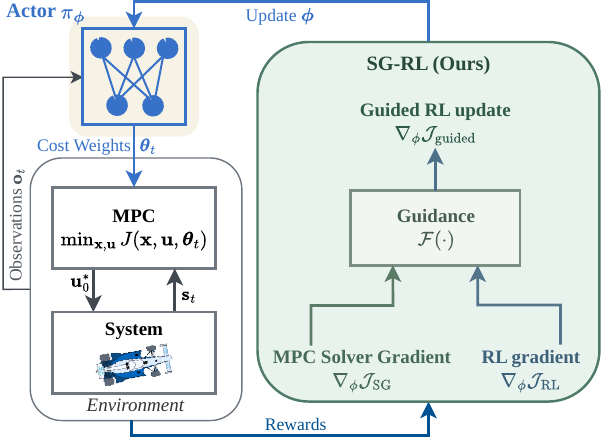}
    \caption{\gls{sgrl} learns an online MPC weight-adaptation policy $\pi_{\phi}$ by combining environment-based policy-gradient learning $\nabla_{\phi}\mathcal{J}_{\mathrm{RL}}$ with model-based solver-gradient guidance $\nabla_{\phi}\mathcal{J}_{\mathrm{SG}}$. The update preserves closed-loop return optimization while using solver sensitivities as bounded, model-informed guidance for improved sample efficiency and performance.}
    \label{fig:sgrl_teaser}
\end{figure}

\gls{mpc} is a powerful framework for constrained optimal control of complex dynamical systems~\cite{rawlings_model_2017}. Its closed-loop performance depends heavily on the formulation of the optimization problem, including the system model, constraints, and cost function. In many applications, the model and constraints are fixed to preserve safety guarantees. The cost function remains a primary mechanism for shaping controller closed-loop behavior and trading off competing objectives~\cite{romero_actor_critic_2024}. 
However, a single fixed set of cost-function weights is often insufficient when operating conditions change. Different scenarios may require different trade-offs between objectives, making the relationship between cost weights and closed-loop performance highly context dependent. This observation has motivated the development of \emph{Weights-varying \gls{mpc}} approaches~\cite{Zarrouki2021_weights_varying,zarrouki2024_safe_rl,2025GB_PL_DiffNMPC}, which adapt the cost-function weights online while keeping the underlying model and constraints unchanged. Such adaptation is particularly relevant in safety-critical and highly dynamic applications. A prominent example is the combined nonlinear longitudinal-lateral control problem in autonomous vehicle racing~\cite{betz2022autonomous}, where operating near the handling limits requires different performance trade-offs across racetrack segments.

Learning such policies is challenging because it induces a bilevel optimization structure~\cite{franceschi2018bilevel}. Specifically, the return depends on the policy parameters only indirectly, through the numerical solution of a constrained \gls{mpc}; decisions at one time step affect future states and therefore future \gls{mpc} solutions, resulting in long-horizon credit-assignment problems; the \gls{mpc} solver introduces implicit, nonlinear, and potentially non-smooth mappings that complicate gradient computation; and the policy must be learned in closed-loop interaction with the controlled system.

Existing approaches for MPC cost-weight optimization can broadly be divided into two categories. The first computes a single fixed parameter vector offline, whereas the second learns a state-dependent weight-adaptation policy online. Classical offline tuning methods, such as genetic algorithms and Bayesian optimization~\cite{ramasamy2019optimal,rodrigues2019tuning,sorourifar2021data,rupenyan2021performance}, can identify useful fixed weights but do not address online adaptation to changing conditions. Since this work addresses online adaptation, we focus on the latter category. In the taxonomy of~\cite{reiter_2026_synthesis}, this corresponds to a hierarchical \gls{rl}--MPC architecture in which the MPC is part of the actor and a neural policy supplies the MPC parameters. Within this setting, recent work distinguishes two complementary sources of learning signal: methods that build \textbf{\emph{environment-based gradients}} from sampled closed-loop interactions (black-box learning), and methods that compute \textbf{\emph{model-based solver gradients}} by differentiating through the MPC solver (white-box learning).

The first family learns online cost-weight adaptation policies using \emph{environment-based gradients}. The environment-based baselines considered here construct policy updates from sampled closed-loop returns, treating the MPC controller as part of the environment~\cite{Zarrouki2021_weights_varying,zarrouki2024_safe_rl,reiter_2023_hierarchical,Bohn2023}. \gls{rl} has also been applied more broadly to learning fixed MPC parameters, such as terminal costs or prediction horizons~\cite{bohn2021reinforcement,Bohn2023}, while the concept of \emph{Weights-varying MPC} introduced learning state-dependent cost-weight adaptation policies~\cite{Zarrouki2021_weights_varying,zarrouki2024_safe_rl}. More generally, MPC has been incorporated as a structured policy or value-function approximator within actor-critic architectures, deterministic and stochastic policy-gradient methods, economic MPC, learning-based MPC formulations, and joint system-identification and control frameworks~\cite{zanon2019practical,gros_data_2020,gros2020mixed_integer,gros2021stochastic_pg,kordabad2021battery,seel2023qlearning_dpg,martinsen2020sysid_rl_mpc,zarrouki_adaptive_2024,sawant2023bigdata}. These baselines benefit from long-horizon credit assignment and can use reward signals defined by broad classes of closed-loop objectives, including non-smooth or sparse terms. However, their updates still rely on sampled closed-loop interactions and inherit the usual approximation, finite-horizon, and estimator-bias issues of practical policy-gradient methods.

The second family learns MPC parameters using \emph{model-based solver gradients} obtained from differentiable optimization. Building on task-based learning through optimization problems~\cite{donti2017task}, differentiable optimization layers~\cite{amos2017optnet,agrawal2019differentiable}, and differentiable MPC~\cite{amos_differentiable_2019}, recent advances enable efficient backpropagation through increasingly general MPC formulations, including constrained nonlinear MPC~\cite{frey_differentiable_2025,fichtner2025leapc}. Until recently, such approaches focused almost exclusively on optimizing fixed MPC parameters through solver sensitivities, including DiffTune-MPC~\cite{tao_difftune_mpc_2024}, differentiable actor learning~\cite{romero2025actor}, cost-matching approaches~\cite{cai2026costmatching}, and ZipMPC~\cite{rickenbach_zipmpc_2025}. Only very recently, gradient-based policy learning (\gls{gbpl}) for nonlinear MPC~\cite{2025GB_PL_DiffNMPC} extended the differentiable nonlinear MPC to learn online cost-weight adaptation policies. Our \gls{gbpl} baseline is built upon~\cite{2025GB_PL_DiffNMPC} and uses a recent backpropagation software feature in \texttt{acados}~\cite{frey_differentiable_2025} to efficiently compute solver gradients for nonlinear constrained \gls{mpc}. In this view, \gls{gbpl} plays a cost-matching role: it aligns a differentiable MPC surrogate loss with task-level performance terms on trajectories generated by the predictive model. Unlike environment-based \gls{rl} updates, these methods exploit local first-order gradients of the differentiable surrogate under solver-regularity assumptions, resulting in substantially higher sample efficiency when the surrogate is informative.

Despite their sample efficiency, differentiable-solver methods optimize model-based surrogate objectives whose gradients are only as accurate as the predictive model and therefore may become misleading under model mismatch. Furthermore, they generally lack the exploration mechanisms and long-horizon value estimation that make \gls{rl} effective for optimizing the performance of the realized closed-loop. Conversely, our environment-based PPO baseline does not use the rich first-order information available from differentiable MPC solvers. The methods compared in this work therefore emphasize one of two complementary information sources: either sampled closed-loop returns or solver sensitivities. Environment-based gradients more directly reflect the realized closed-loop objective but can suffer from high variance, whereas solver gradients provide lower-variance local guidance but inherit bias from the predictive model. Their complementary properties motivate the combination of both within a single learning framework. This gap motivates the central question of this paper: how can environment-based and model-based solver gradients be combined effectively for learning online MPC cost-weight adaptation policies?

This question is related to a broader family of guided, structured, and hybrid reinforcement-learning methods that combine model-free policy updates with model-based gradients, planning modules, auxiliary losses, or teacher signals~\cite{levine2014guided,heess2015learning,tamar2016value,parmas2018pipps,clavera2020model,xu2022shac,hansen2022tdmpc,hansen2024tdmpc2,Schmitt2018}. These methods demonstrate that combining complementary gradient estimators can substantially improve learning efficiency. However, these works do not use differentiable constrained nonlinear MPC solver sensitivities as structured auxiliary gradients for policy optimization over online MPC cost weights.

Within the differentiable MPC literature, solver sensitivities have also been used beyond direct parameter optimization. Adhau et al.~\cite{ADHAU202311841} employ nonlinear-program sensitivities to approximate action-value quantities in a Q-learning framework, thereby reducing the number of MPC evaluations required during learning. This approach, however, neither learns an online cost-weight adaptation policy nor uses solver sensitivities to guide policy optimization itself. Romero et al.~\cite{romero2025actor} propagate solver gradients through a differentiable linear MPC layer embedded in a PPO actor for quadrotor control, while related hybrid schemes study differentiable simulation gradients~\cite{suh2022differentiable} or use \gls{rl} to provide NMPC reference guidance~\cite{Reiter2026AC4MPCActorC}. In contrast, our differentiable nonlinear MPC-based work does not replace environment-based policy gradients with model-based solver gradients, nor does it guide the controller only through references. Instead of viewing solver gradients as an alternative optimization objective, we use them as auxiliary signals inside PPO. Consequently, the optimization objective remains the sampled closed-loop return, while solver sensitivities provide local model-informed search directions for online cost-weight adaptation with constrained nonlinear MPC. 

The contributions of this work are threefold:

\begin{compactenum}

\item \textbf{Solver-gradient Guide for online MPC weight adaptation.}
We introduce Solver-Gradient Guided Reinforcement Learning (\gls{sgrl}), a framework for online cost-weight adaptation in constrained nonlinear \gls{mpc}. Compared to \gls{rl}, which learns from environment-based policy gradients estimated from sampled closed-loop returns, and differentiable-solver approaches that optimize model-based surrogate objectives using solver sensitivities, \gls{sgrl} uses solver-gradient guidance as an auxiliary signal for \gls{rl} while still optimizing the realized closed-loop return (Sec.~\ref{sec:sgd_rl}).

\item \textbf{Four methods for guiding \gls{ppo} through solver sensitivities,} We instantiate this mechanism through four modular \gls{sgrl} algorithms that integrate the same solver-gradient guidance into distinct PPO components: actor-update scaling, performance-gated auxiliary policy learning, solver-guided advantage estimation, and solver-aware critic learning. These methods provide four complementary instantiations of the \gls{sgrl} framework and are evaluated separately to understand which forms of solver guidance are most effective in practice (Sec.~\ref{sec:sgd_rl}).


\item \textbf{Full-scale validation under model mismatch.}
We demonstrate our \gls{sgrl} on a closed-loop high-dimensional, highly nonlinear control task: nonlinear \gls{mpc} for combined longitudinal and lateral vehicle control on two full-scale, high-fidelity autonomous racing simulation platforms, following racelines near the vehicle handling limits under intentional model mismatch. In this setting, all \gls{sgrl} variants outperform the PPO baseline in peak closed-loop performance, reach PPO's best return with up to 70.6\,\% fewer training samples, and outperform the \gls{gbpl} baselines by at least 54\,\% in return, while also demonstrating zero-shot generalization and interpretable weight-adaptation strategies (Sec.~\ref{sec:application_AV}-\ref{sec:results}).

\end{compactenum}

\vspace{0.3em}
\noindent\textbf{Notation and Background.}
\label{sec:ppo_background}
We denote by $\pi_\phi(\cdot \mid \boldsymbol{o}_t)$ a stochastic actor that maps the observation $\boldsymbol{o}_t$ to a distribution over MPC weights, with mean $\boldsymbol{\mu}_\phi(\boldsymbol{o}_t)$, while  $V_\psi(\boldsymbol{o}_t)$ denotes the critic and provides the value function. We use the generalized-advantage estimate $\hat{A}_t$~\cite{schulman2015gae} to quantify whether the sampled weight vector performed better or worse than expected. With TD residual $\delta_t := r_t + \gamma V_\psi(\boldsymbol{o}_{t+1}) - V_\psi(\boldsymbol{o}_t)$, we write
\begin{equation}
    \hat{A}_t = \delta_t + \gamma\lambda(1-d_{t+1})\hat{A}_{t+1},
\end{equation}
where $\gamma$ is the discount factor and $\lambda$ is the GAE parameter~\cite{schulman2015gae}. \glsentryshort{ppo} updates the actor through the clipped surrogate objective
\begin{equation} \label{eq:ppo_surrogate}
\begin{aligned}
    \mathcal{J}_\mathrm{PPO}(\boldsymbol{\phi}) = \hat{\mathbb{E}}_t \Big[
    \min\big(&\rho_t(\boldsymbol{\phi}) \hat{A}_t, \\
             &\text{clip}\big(\rho_t(\boldsymbol{\phi}), 1-\epsilon, 1+\epsilon\big) \hat{A}_t \big) 
    \Big],
\end{aligned}
\end{equation}
with sampled action $\boldsymbol{a}_t\equiv\boldsymbol{\theta}_t$, ratio $\rho_t(\boldsymbol{\phi}) = \pi_\phi(\boldsymbol{a}_t \mid \boldsymbol{o}_t)/\pi_{\phi_{\mathrm{old}}}(\boldsymbol{a}_t \mid \boldsymbol{o}_t)$, and clipping range $\epsilon$. For the critic, we use the standard value-regression term based on the fixed target $\hat{R}^{\lambda}_t := \hat{A}_t + V_{\boldsymbol{\psi}_{\mathrm{old}}}(\boldsymbol{o}_t)$.\\
Actor and critic are then optimized through the standard combined \glsentryshort{ppo} loss
\begin{equation}
\label{eq:PPO_loss}
\begin{aligned}
    \mathcal{L}_\mathrm{PPO}(\boldsymbol{\phi},\boldsymbol{\psi}) = &\; -\mathcal{J}_\mathrm{PPO}(\boldsymbol{\phi}) + c_V\,\mathcal{L}_V(\boldsymbol{\psi}) \\
    &\; - c_{\mathrm{ent}}\,\hat{\mathbb{E}}_t\big[\mathcal{H}(\pi_\phi(\cdot\,|\,\boldsymbol{o}_t))\big],
\end{aligned}
\end{equation}
where $c_V$ and $c_{\mathrm{ent}}$ weight the value and entropy terms, respectively. The standard \gls{rl} update used as baseline later with learning rate $\eta$ is
\begin{equation} \label{eq:rl_update_std}
    \mathcal{U}_\mathrm{\gls{rl}}: \; (\boldsymbol{\phi}_{k+1},\boldsymbol{\psi}_{k+1}) = (\boldsymbol{\phi}_k,\boldsymbol{\psi}_k) - \eta\,\nabla_{\phi,\psi}\,\mathcal{L}_\mathrm{PPO}(\boldsymbol{\phi},\boldsymbol{\psi})
\end{equation}
\section{Problem Formulation}
\label{sec:problem_formulation}

We formalize weights-varying \gls{nmpc} as a hierarchical decision-making problem (Fig.~\ref{fig:sgrl_teaser}). During deployment, a low-level \gls{nmpc} controller solves a constrained finite-horizon optimal control problem at each sampling instant, while a high-level learning policy adapts the \gls{nmpc} cost weights online along the task domain, thereby providing a mechanism to shape long-horizon behavior. During learning, the same closed-loop architecture is rolled out to collect trajectories, and an outer optimization loop updates the policy parameters across training iterations. 
\subsection{Weights-varying Nonlinear MPC}
\subsubsection{Motivation: Why Adapt Cost Weights?}

Standard \gls{nmpc} typically relies on fixed cost weights tuned offline to balance competing objectives such as tracking accuracy and control effort. In practice, the best trade-off is context-dependent: operating regime, disturbance level, uncertainty, and anticipated task difficulty change over time, so a single static tuning is inevitably conservative in some conditions and too aggressive in others. We therefore adapt the cost weights online, which reshapes the optimization landscape and the relative emphasis placed on these objectives without modifying the underlying model or constraints.

\subsubsection{Weights-varying \gls{nmpc} OCP Formulation}

The OCP solved at time $t$ with a receding prediction horizon $N$ is:
\begin{mini!}[short]
    {\mathbf{x}, \mathbf{u}}
    {J(\mathbf{x}, \mathbf{u}, \boldsymbol{\theta}_t) = \sum_{k=0}^{N-1} l(\mathbf{x}_k, \mathbf{u}_k, \boldsymbol{\theta}_t) + l_N(\mathbf{x}_N, \boldsymbol{\theta}_t)}
    {\label{eq:nmpc_ocp}}
    {}
    \addConstraint{\mathbf{x}_0}{= \mathbf{s}_t}
    \addConstraint{\mathbf{x}_{k+1}}{= \boldsymbol{f}(\mathbf{x}_k, \mathbf{u}_k),}{\qquad k=0, \dots, N-1 \label{eq:NMPC_system_dyna}}
    \addConstraint{0}{\leq \boldsymbol{h}(\mathbf{x}_k, \mathbf{u}_k),}{\qquad k=0, \dots, N-1 \label{eq:NMPC_constr}}
    \addConstraint{0}{\leq \boldsymbol{h}_N(\mathbf{x}_N)}{\label{eq:NMPC_constr_final}}
\end{mini!}

where $\mathbf{x} = [\mathbf{x}_0, \dots, \mathbf{x}_N]$ and $\mathbf{u} = [\mathbf{u}_0, \dots, \mathbf{u}_{N-1}]$ are the state and control trajectories, while $\mathbf{s}_t$ denotes the current state estimate available to the controller at time $t$. The stage cost $l(\cdot)$ and the terminal cost $l_N(\cdot)$ are parameterized by a weight vector $\boldsymbol{\theta}_t \in \mathbb{R}^{n_\theta}$, which is fixed on the prediction horizon at time $t$ but may change between sampling instants. The function $\boldsymbol{f}(\cdot)$ in \eqref{eq:NMPC_system_dyna} is the internal prediction model of the controller. In this work, we keep $\boldsymbol{f}(\cdot)$ fixed and adapt only the cost parameters. The functions \eqref{eq:NMPC_constr}-\eqref{eq:NMPC_constr_final} provide inequality constraints on the states and controls.
\subsubsection{Learning-based Cost-Weight Adaptation Policies}
We introduce a high-level policy $\pi_\phi$ (parameterized by $\boldsymbol{\phi}$) that maps an observation $\boldsymbol{o}_t$ to a distribution over \gls{nmpc} weight vectors (Fig.~\ref{fig:sgrl_teaser}). The applied weight vector is denoted $\boldsymbol{\theta}_t$ and is equivalent to the RL action $\boldsymbol{a}_t$, while $\boldsymbol{\mu}_\phi(\boldsymbol{o}_t)$ denotes the policy mean used for deterministic evaluation and solver-guided policy updates. The policy is evaluated online both during training rollouts and during deployment; the difference between the two phases is only whether $\boldsymbol{\phi}$ is updated. 

At each time $t$, the lower-level \gls{nmpc} solves \eqref{eq:nmpc_ocp} with the current state $\mathbf{s}_t$ and the set of parameters $\boldsymbol{\theta}_t$, then applies the first input $\mathbf{u}_t := \mathbf{u}^*_0(\mathbf{s}_t, \boldsymbol{\theta}_t)$ to the system in a receding-horizon fashion.

\subsection{Problem Statement: Learning Policies for Weights-varying \gls{nmpc}} \label{sec:learning_policies}
Our objective is to learn the parameters $\boldsymbol{\phi}$ of the high-level weight-selection policy to optimize long-horizon closed-loop performance. Let $\mathcal{L}_{\mathrm{perf}}(\mathbf{s}_t, \mathbf{u}_t)$ denote a user-defined \emph{instantaneous} performance loss evaluated on the realized closed-loop state and the control actually applied at time $t$ (Fig.~\ref{fig:bilevel_architecture}). This high-level loss can differ from the \gls{nmpc} internal cost $J$ and may include objectives that are difficult to encode directly in the \gls{nmpc}, including non-smooth, sparse, or history-dependent terms such as collisions, failure-to-finish, or penalties on rapid weight variation.

The objective is to find the optimal policy parameters $\boldsymbol{\phi}^*$ that minimize the expected cumulative loss over a closed-loop episode of length $T$, where the expectation is taken over the trajectories induced by the policy, the \gls{nmpc} controller, the true rollout environment, and exogenous stochasticity:
\begin{equation}
    \boldsymbol{\phi}^* = \argmin_\phi \; \mathbb{E}_{\tau \sim p^{\mathrm{cl}}_{\phi}} \left[ \sum_{t=0}^{T} \mathcal{L}_{\mathrm{perf}}(\mathbf{s}_t, \mathbf{u}_t) \right],
    \label{eq:closed_loop_learning_objective}
\end{equation}

Learning the high-level weight-selection policy is challenging because the return depends on $\boldsymbol{\phi}$ only indirectly through the closed-loop interaction of the policy $\pi_\phi$, the numerical solution of a constrained \gls{nmpc} problem at each time step, and the true plant or rollout-environment dynamics. Decisions at time $t$ therefore affect future states, observations, and solver calls, which induces long-horizon credit assignment during training.

The following sections develop three approaches for solving the upper-level optimization problem in \eqref{eq:closed_loop_learning_objective}. We first introduce the neural policy architecture used throughout the paper (Sec.~\ref{sec:nn_policy}), then present two baseline optimization paradigms based on differentiable optimization (Sec.~\ref{sec:gb_pl}) and reinforcement learning (Sec.~\ref{sec:rl_approach}), before introducing our proposed Solver-Gradient Guided Reinforcement Learning framework (Sec.~\ref{sec:sgd_rl}), which combines the strengths of both.


\section{Designing a Neural Policy for Weights-varying \gls{nmpc}}
\label{sec:nn_policy}

Weights-varying \gls{nmpc} depends on a policy $\pi_\phi$ that maps high-dimensional observations to effective cost configurations. This mapping is implicit, nonlinear, and shaped by \gls{nmpc} constraints, so fixed offline weights are insufficient when task requirements and disturbances change quickly.

We design a neural \textit{meta-controller} in the \gls{nmpc} hyper-parameter space. At each step, it maps $\boldsymbol{o}_t$ to weights $\boldsymbol{\theta}_t$ (Figs.~\ref{fig:sgrl_teaser},~\ref{fig:rwp_architecture}) for the next planning horizon. This hierarchical decomposition allows the \gls{nmpc} to handle system constraints and low-level dynamics, while the neural policy focuses on shaping the broader behavioral objectives.

\subsection{Policy Architecture: History-Aware MLP} \label{sec:policy_arch}
\begin{figure}[!t]
    \centering
    \includegraphics[width=0.95\linewidth]{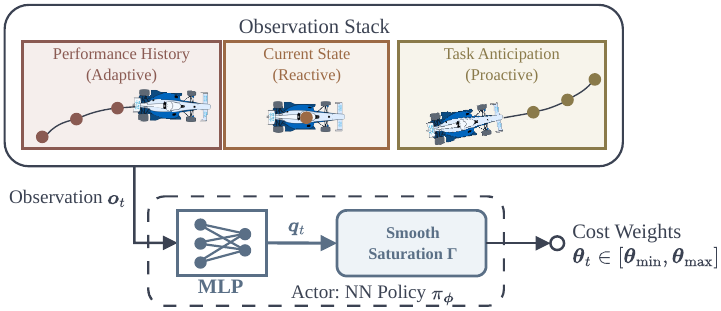}
    \caption{Architecture of the online cost-weight policy. 
    }
    \label{fig:rwp_architecture}
\end{figure}

The main design choices are the policy input $\boldsymbol{o}_t$ and output $\boldsymbol{\theta}_t$: the input must support reaction, anticipation, and mismatch adaptation, while the output is restricted to valid \gls{nmpc} weight ranges. This reduces the learning search space and improves robustness.
We parameterize our policy $\pi_\phi$ as a feed-forward neural network. 
The policy and \gls{nmpc} share the online sensing/estimation pipeline: \gls{nmpc} initializes the OCP with $\mathbf{s}_t$, and the policy receives features $\boldsymbol{o}_t$ from measured/estimated quantities, reference previews, and recent performance.
To capture temporal context without the complexity of Recurrent Neural Networks (RNNs), we explicitly structure $\boldsymbol{o}_t$ into three functional groups (Fig.~\ref{fig:rwp_architecture}):
\begin{compactenum}
    \item \textbf{Current measurable context (Reactive):} task-relevant local state information in a task-relative frame, enabling reactive decisions.
    \item \textbf{Reference preview / task context (Proactive):} a short-horizon lookahead $\mathbf{r}_{t:t+H_\text{ant}}$ of the reference quantities used by the \gls{nmpc}, such as target speed and path geometry, so weights change before braking, curvature, or acceleration events.
    \item \textbf{Performance history (Adaptive):} recent closed-loop tracking terms, e.g., squared lateral and velocity errors, summarizing whether the current schedule works and helping infer persistent disturbances, model mismatch, or actuator transients.
\end{compactenum}

\subsection{Action Space: Constrained Cost Parameterization}
\label{subsec:action_space}

Our policy outputs cost weights $\boldsymbol{\theta}_t$, which must remain positive and well scaled. Arbitrary weights can destroy objective conditioning or strict convexity for diagonal quadratic terms, yield unbounded solutions, or drive training into numerically unstable magnitudes.

Instead of a matrix-level projection onto the positive-definite cone, we use a \textit{Constrained Cost Parameterization}: each weight is bounded within strictly positive limits. These conservative box constraints are simple, robust, and enforce positive-definiteness for diagonal costs. Domain knowledge sets $[\boldsymbol{\theta}_{\min}, \boldsymbol{\theta}_{\max}]$, and unconstrained logits $\mathbf{q}_t \in \mathbb{R}^{n_\theta}$ are mapped to this range by a scaled \texttt{tanh}:
\begin{equation}
    \boldsymbol{\theta}_t = \boldsymbol{\Gamma}(\mathbf{q}_t) = \frac{\boldsymbol{\theta}_{\max} - \boldsymbol{\theta}_{\min}}{2} \tanh(\mathbf{q}_t) + \frac{\boldsymbol{\theta}_{\max} + \boldsymbol{\theta}_{\min}}{2}.
\end{equation}
This mechanism serves two purposes:
\begin{compactenum}
    \item \textbf{Search-space regularization:} confines optimization to a bounded hyper-rectangle, preventing divergence and removing irrelevant parameter regions.
    \item \textbf{Smooth bounded mapping:} keeps the parameterization differentiable and avoids hard clipping, at the cost of attenuated gradients near boundaries.
\end{compactenum}

\begin{table}[!t]
\caption{Comparison of Policy Architectures for Weights-varying \gls{nmpc}. 
Our proposed policy uniquely combines reactivity, adaptability, and proactive anticipation within a constrained action space.
}
\label{tab:policy_comparison}
\centering
\renewcommand{\arraystretch}{1.1}
\setlength{\tabcolsep}{3pt}
\footnotesize
\begin{tabular}{@{}m{1.5cm} c c c m{3.0cm}@{}}
\toprule
\textbf{Method}  & \textbf{Reactive} & \textbf{Adaptive} & \textbf{Proactive} & \textbf{Action Space} \\
\midrule
\cite{Zarrouki2021_weights_varying} & $\checkmark$ & $\checkmark$ & $\times$ & Bounded (Sigmoid) \\
\hline
\cite{zarrouki2024_safe_rl}  & $\checkmark$ & $\checkmark$ & $\checkmark$ & Bounded, discrete \\
\hline
\cite{2025GB_PL_DiffNMPC}  & $\times$ & $\times$ & $\checkmark$ & Unbounded (Softplus) \\
\hline
\textbf{Ours}  & $\checkmark$ & $\checkmark$ & $\checkmark$ & \textbf{Bounded (Tanh)} \\
\bottomrule
\end{tabular}
\end{table}



\section{\gls{gbpl} for Weights-varying Differentiable \gls{nmpc}}
\label{sec:gb_pl}
This section presents the \gls{gbpl} baseline for weights-varying differentiable \gls{nmpc}. Here, \gls{gbpl} refers to the setting in which a differentiable surrogate loss is backpropagated through the \gls{nmpc} layer and the neural policy. We use both the prior reduced \gls{gbpl} formulation \cite{2025GB_PL_DiffNMPC} and our new extended implementation as baselines for evaluating the \gls{sgrl} methods introduced later in Section~\ref{sec:sgd_rl}.

\subsection{Differentiating the \gls{nmpc} Solution}
To propagate gradients through the \gls{nmpc} solver, we follow the sensitivity analysis of nonlinear programs in \cite{frey_differentiable_2025}. Let $\mathbf{w}^*(\boldsymbol{\theta}_t) = [\mathbf{z}^*, \boldsymbol{\lambda}^*, \boldsymbol{\mu}^*]$ denote the optimal primal-dual solution vector of the OCP \eqref{eq:nmpc_ocp} given weights $\boldsymbol{\theta}_t$, where $\mathbf{z}^*$ contains the optimal states and controls, and $\boldsymbol{\lambda}^*, \boldsymbol{\mu}^*$ are the Lagrange multipliers associated with the equality and inequality constraints. The optimality of $\mathbf{w}^*$ is defined by the Karush-Kuhn-Tucker (KKT) conditions, which can be expressed as a system of nonlinear equations $\boldsymbol{R}(\mathbf{w}^*, \boldsymbol{\theta}_t) = 0$.
Assuming the standard regularity conditions (Linear Independence Constraint Qualification, Second-Order Sufficiency Condition, and Strict Complementarity) hold locally, the Implicit Function Theorem (IFT) yields a locally unique solution map whose derivative exists piecewise, i.e., within regions where the active set remains unchanged. The sensitivity matrix $\mathbf{S}_t$, representing the Jacobian of the solution with respect to the parameters, then follows from differentiating the KKT residual equations with respect to $\boldsymbol{\theta}_t$:
\begin{equation}
    \label{eq:NMPC_sensit}
    \mathbf{S}_t = \frac{\partial \mathbf{w}^*}{\partial \boldsymbol{\theta}_t} = - \left[ \frac{\partial \boldsymbol{R}}{\partial \mathbf{w}^*} \right]^{-1} \frac{\partial \boldsymbol{R}}{\partial \boldsymbol{\theta}_t}.
\end{equation}
In this work, we employ the recent differentiable-solver sensitivity functionality implemented in \texttt{acados}~\cite{verschueren_acadosmodular_2022, frey_differentiable_2025}. Concretely, the solver returns forward sensitivities of the NLP solution with respect to the cost weights by solving the linearized KKT system associated with the current SQP iterate. This yields the Jacobian $\nabla_{\boldsymbol{\theta}_t} \mathbf{z}^*$, quantifying how the optimal trajectory changes with infinitesimal perturbations in the cost weights.
\subsection{Gradient Propagation and Policy Learning} \label{sec:loss_func_grad}
The goal of \gls{gbpl} is to update the policy-parameter collection $\boldsymbol{\phi}$, comprising all trainable weights and biases of the neural policy, by differentiating through the \gls{nmpc} solver. Since this requires end-to-end differentiability, \gls{gbpl} optimizes a \emph{differentiable surrogate} loss $\tilde{\mathcal{L}}_{\text{perf}}(\mathbf{z}^*)$ defined on the \gls{nmpc} solution $\mathbf{z}^*$ (predicted state/control trajectory) returned by the solver.
This surrogate is a differentiable proxy for the task-level loss $\mathcal{L}_{\text{perf}}(\cdot)$ introduced in Section~\ref{sec:learning_policies}. In our setting, it is built by reusing the same smooth tracking, velocity, and control-regularization terms as the task-level loss, but evaluating them on the \gls{nmpc}-predicted open-loop trajectory $\mathbf{z}^*$ rather than the realized closed-loop trajectory. Consequently, \gls{gbpl} is restricted to objectives for which such a smooth surrogate can be written; genuinely non-smooth or sparse task terms must either be approximated by smooth surrogates or omitted from the \gls{gbpl} update, which is itself a limitation of the approach rather than something resolved here.

We compute the corresponding model-based end-to-end gradient of $\tilde{\mathcal{L}}_{\text{perf}}$ with respect to the policy parameters $\boldsymbol{\phi}$ at time $t$ using the chain rule:
\begin{equation} \label{eq:chain_rule}
    \nabla_\phi \tilde{\mathcal{L}}_\text{perf} = \underbrace{\left(\underbrace{\frac{\partial \tilde{\mathcal{L}}_\text{perf}}{\partial \mathbf{z}^*}}_{\text{Loss Gradient}} \cdot \underbrace{\frac{\partial \mathbf{z}^*}{\partial \boldsymbol{\theta}_t}}_{\text{Solver Sensitivity}}\right)}_{:=\,\mathbf{g}^{\theta}_{\mathrm{SG},t}} \cdot \underbrace{\frac{\partial \boldsymbol{\theta}_t}{\partial \boldsymbol{\phi}}}_{\text{Policy Jacobian}}
\end{equation}
\begin{compactenum}
    \item \textbf{Loss Gradient:} $\frac{\partial \tilde{\mathcal{L}}_\text{perf}}{\partial \mathbf{z}^*}$ is computed using automatic differentiation (e.g., PyTorch). It represents the sensitivity of the differentiable surrogate loss with respect to the optimal trajectory produced by the \gls{nmpc}.
    \item \textbf{Solver Sensitivity:} $\frac{\partial \mathbf{z}^*}{\partial \boldsymbol{\theta}_t}$ is extracted from the sensitivity matrix $\mathbf{S}_t$ in \eqref{eq:NMPC_sensit}, computed by \texttt{acados}. This step bridges the optimization landscape with the control parameters.
    \item \textbf{Weight-Space Solver Gradient:} The product $\mathbf{g}^{\theta}_{\mathrm{SG},t}$ is the local gradient of the surrogate loss in \gls{nmpc}-weight space. The corresponding solver-informed descent step is $-\mathbf{g}^{\theta}_{\mathrm{SG},t}$. We reuse this notation in Section~\ref{sec:sgd_rl}, where the same solver-derived loss-gradient direction is normalized and injected into PPO.
    \item \textbf{Policy Jacobian:} $\frac{\partial \boldsymbol{\theta}_t}{\partial \boldsymbol{\phi}}$ maps the weight-space solver loss gradient back to the trainable parameters of the neural policy.
\end{compactenum}
Thus, \gls{gbpl} computes an exact model-based gradient of the surrogate loss $\tilde{\mathcal{L}}_\text{perf}(\mathbf{z}^*)$ by differentiating through the solver and the policy.

Formally, the update rule of \gls{gbpl} at iteration $k$, over a data batch $\mathcal{B}$ with learning rate $\eta$, is:
\begin{equation} \label{eq:gb_update}
    \mathcal{U}_\mathrm{GB}: \boldsymbol{\phi}_{k+1} = \boldsymbol{\phi}_k - \eta \frac{1}{|\mathcal{B}|} \sum_{i \in \mathcal{B}} \nabla_\phi \tilde{\mathcal{L}}_\text{perf}^{(i)}.
\end{equation}

\subsubsection*{Key Novelties} Although pure \gls{gbpl} is not our main focus, we introduce three structural improvements to the existing reduced \gls{gbpl} formulation in \cite{2025GB_PL_DiffNMPC}. 
First, instead of updating the policy after each rollout in the order it is generated, we collect a batch of rollouts, shuffle them into mini-batches, and perform multiple optimization passes (epochs) over this batch before collecting new data. This improves performance and sample efficiency. 
Second, we constrain the policy output using a differentiable $\tanh$ projection (Section~\ref{subsec:action_space}), ensuring all weights remain within a safe, bounded region and avoiding the poorly scaled objectives that can arise with unbounded parameterizations. Third, our observation space is history-aware, combining temporal stacks of current states and tracking errors with future references, which enables the policy to adapt reactively to disturbances and persistent errors, rather than relying solely on feedforward prediction. These changes collectively yield a more stable, adaptive, and scalable \gls{gbpl} framework for weights-varying differentiable \gls{nmpc}.

\subsection{\gls{gbpl} Algorithm Limitations}
While \gls{gbpl} provides a direct path for optimizing a differentiable, model-based surrogate loss $\tilde{\mathcal{L}}_{\text{perf}}(\cdot)$, it suffers from specific structural limitations when the goal is closed-loop performance optimization:
\begin{compactenum}
    \item \textbf{Local descent without exploration:} Like other local gradient methods, \gls{gbpl} follows the immediate descent direction of its surrogate objective and does not include an explicit exploration mechanism. Its training outcome is therefore sensitive to initialization and to the local geometry of the surrogate landscape.
    \item \textbf{Need for differentiable loss:} \gls{gbpl} requires the surrogate loss $\tilde{\mathcal{L}}_\text{perf}(\mathbf{z}^*)$ to be differentiable with respect to the \gls{nmpc} trajectory $\mathbf{z}^*$. This precludes optimizing sparse event-based objectives (e.g., "Lap completed", "Collision") or using non-smooth penalties.
    \item \textbf{Open-loop, model-dependent tuning:} The sensitivities in \eqref{eq:chain_rule} are derived from the \gls{nmpc}'s internal open-loop predictions rather than from realized closed-loop trajectories. As a result, \gls{gbpl} primarily tunes the MPC cost to match the task loss under the predictive model used by the solver, rather than directly optimizing the true closed-loop behavior of the plant. This cost-matching view is closely related to recent model-based MPC cost-matching formulations \cite{cai2026costmatching}. When the predictive model is inaccurate, the resulting gradient can systematically deviate from the true closed-loop improvement direction.
\end{compactenum}

\section{Reinforcement Learning Baseline for Weights-varying \gls{nmpc}}
\label{sec:rl_approach}

To address the limitations of \gls{gbpl}—specifically the myopic nature of local gradient descent and the requirement of differentiable surrogate losses—we formulate \gls{nmpc} weight adaptation within a \gls{rl} framework and use it as the second benchmark for solver-guided RL (Sec.~\ref{sec:sgd_rl}). 
Our baseline is PPO, an on-policy actor--critic method that updates the stochastic weight policy $\pi_\phi$ from sampled closed-loop returns without solver differentiation.
Building on RL-WMPC~\cite{Zarrouki2021_weights_varying,zarrouki2024_safe_rl}, the plant plus \gls{nmpc} controller defines the environment, and the agent adapts \gls{nmpc} weights through interaction. We choose PPO over SAC because its on-policy structure uses MPC real-time iterations more efficiently than random off-policy sampling.

We model the problem as a \gls{pomdp} $\mathcal{M}=(\mathcal{S},\mathcal{A},\mathcal{P},\mathfrak{r},\gamma)$ with latent state space $\mathcal{S}$, action space $\mathcal{A}$, transition kernel $\mathcal{P}$, reward $\mathfrak{r}$, and discount $\gamma$. The action is the \gls{nmpc} weight vector, $\boldsymbol{a}_t\equiv\boldsymbol{\theta}_t$, and $\mathcal{P}$ captures plant--controller closed-loop evolution under those weights. The policy $\pi_\phi(\cdot\mid\boldsymbol{o}_t)$ conditions on observations built from measurements and references (Sec.~\ref{sec:policy_arch}).

To ensure a rigorous comparison against \gls{gbpl}, we define the reward $r_t$ as the negative of the same instantaneous high-level performance loss introduced in Section~\ref{sec:learning_policies}. Thus, the reward is evaluated on the realized closed-loop state and applied input at time $t$, rather than on the internal \gls{nmpc} cost $J$ or on the \gls{nmpc}-predicted trajectory used inside the differentiable surrogate of \gls{gbpl}:
\begin{equation}
    r_t = - \mathcal{L}_{\mathrm{perf}}(\mathbf{s}_{t}, \mathbf{u}^*_0).
\end{equation}
Here $\mathbf{u}^*_0$ denotes the first control input of the \gls{nmpc} solution applied to the system at time $t$. We denote the corresponding expected closed-loop return by
\begin{equation}
    \mathcal{R}(\pi_\phi) := \mathbb{E}_{\pi_\phi}\left[\sum_{t=0}^{T} \gamma^t r_t\right].
\end{equation}
This reward alignment ensures that any performance difference between RL and \gls{gbpl} is attributable to the learning signal and update mechanism rather than to a different task objective. In particular, PPO estimates its policy update from closed-loop returns collected through environment interactions, so the resulting policy gradient is an \textit{environment-based} signal that implicitly reflects exploration, stochasticity, and model mismatch. By contrast, \gls{gbpl}'s update is a \textit{model-based gradient} built from the same underlying performance terms.

The preceding sections reveal a trade-off between differentiable solver-based learning and \gls{rl} for Weights-varying \gls{nmpc}. 
\gls{gbpl} incorporates solver sensitivities, enabling sample-efficient gradient-based policy updates through the \gls{nmpc} layer. However, these updates optimize a differentiable surrogate defined on predicted trajectories and can therefore deviate from the true closed-loop objective under model mismatch or non-smooth task objectives. 
Conversely, \gls{ppo} directly optimizes the expected closed-loop return through environment interaction and can capture model mismatch and long-horizon effects, but can require more interaction data and does not exploit the sensitivity information available from the \gls{nmpc} solver. 
This motivates the central question addressed in the remainder of this paper: 
\textit{how can solver-gradient information from differentiable \gls{nmpc} be incorporated into \gls{rl} to accelerate policy optimization while preserving the closed-loop return as the learning objective?}

\begin{figure*}[!t]
    \centering
    \includegraphics[width=\textwidth]{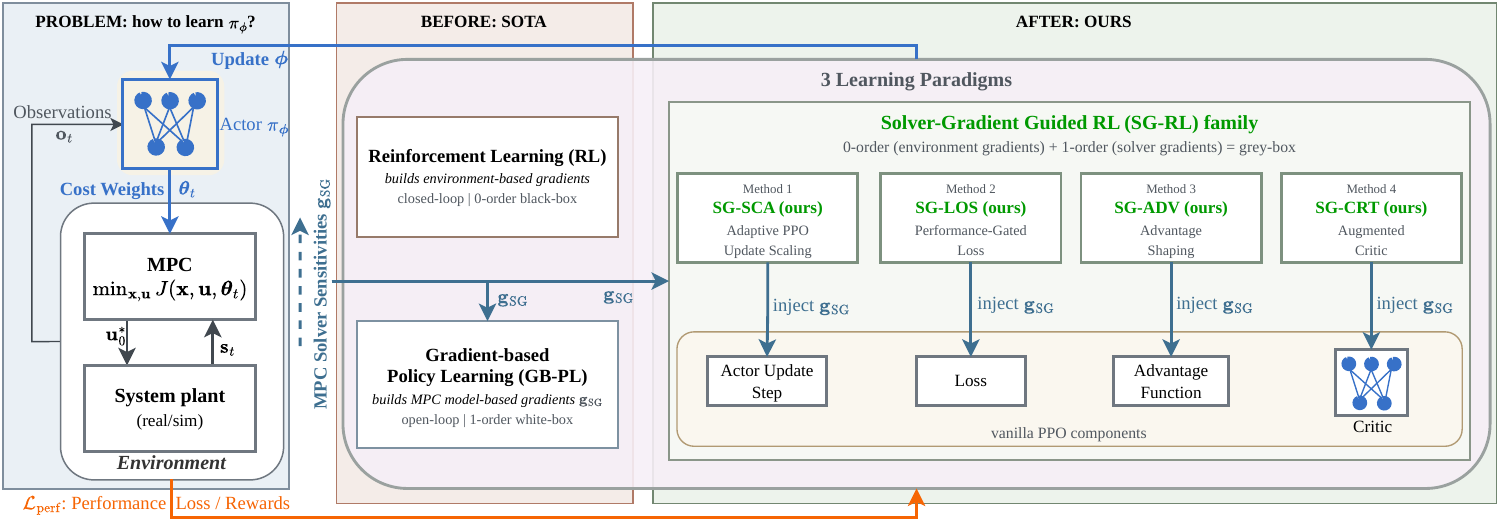}
    \caption{Bi-level Weights-varying \glsentryshort{nmpc} architecture. The policy selects cost weights for the \glsentryshort{nmpc}, while the upper-level learner updates the policy from closed-loop rollouts using sampled performance feedback and solver-gradient guidance.}
    \label{fig:bilevel_architecture}
\end{figure*}
\section{Solver-Gradient Guided Reinforcement Learning}
\label{sec:sgd_rl}

In this section, we introduce our \textbf{Solver-Gradient Guided Reinforcement Learning (\gls{sgrl})}, whose core idea is to inject the model-based \gls{nmpc} solver loss gradient $\mathbf{g}_{\mathrm{SG},t}$ into selected components of the PPO algorithm.

\paragraph{Solver gradients as normalized loss gradients}
Let $\boldsymbol{\theta}_t \in \mathbb{R}^{n_\theta}$ denote the \gls{nmpc} cost-weight vector used by the solver at time $t$. During stochastic rollouts this is the sampled RL action, i.e., $\boldsymbol{a}_t\equiv\boldsymbol{\theta}_t$, whereas deterministic evaluation and the differentiable solver-guidance terms use the policy mean $\boldsymbol{\mu}_\phi(\boldsymbol{o}_t)$ unless stated otherwise. On regular regions of the piecewise-differentiable \gls{nmpc} solution map, we can compute the local sensitivity of the instantaneous performance loss with respect to these weights:
Following Section~\ref{sec:loss_func_grad}, we define the corresponding weight-space solver gradient as the gradient of the differentiable surrogate loss:
\begin{equation}
    \mathbf{g}_{\mathrm{SG},t}^{\theta} := \nabla_{\boldsymbol{\theta}} \, \tilde{\mathcal{L}}_\text{perf}\big(\mathbf{z}^*(\boldsymbol{\theta})\big)\Big|_{\boldsymbol{\theta}=\boldsymbol{\theta}_t}.
\end{equation}

Throughout this section, we use a \emph{direction-only} solver loss-gradient signal by normalizing the sensitivity vector,
\begin{equation}
    \mathbf{g}_{\mathrm{SG},t} := \frac{\mathbf{g}_{\mathrm{SG},t}^{\theta}}{\lVert \mathbf{g}_{\mathrm{SG},t}^{\theta} \rVert_2 + \epsilon},
\end{equation}
so that $\lVert \mathbf{g}_{\mathrm{SG},t} \rVert_2 \le 1$. This removes scale variability across operating contexts and makes the solver-derived signal comparable across time, while leaving update magnitude to the method-specific coefficients that determine how strongly the solver loss-gradient direction is used. A local descent step in weight space is therefore taken along $-\mathbf{g}_{\mathrm{SG},t}$.
When sensitivities are unavailable 
(e.g., infeasible solves), we set $\mathbf{g}_{\mathrm{SG},t}=\mathbf{0}$, recovering standard PPO behavior.

\paragraph{RL gradient convention}
We train PPO by minimizing the compound loss introduced in Eq.~\eqref{eq:PPO_loss}, which combines the clipped policy objective with auxiliary terms such as value regression and entropy regularization. Throughout this section, we denote the objective-level PPO policy gradient by
\begin{equation}
    \mathbf{g}_\mathrm{RL} := \nabla_{\phi} \mathcal{L}_\mathrm{PPO}(\boldsymbol{\phi},\boldsymbol{\psi}).
    \label{eq:g_RL}
\end{equation}
In implementation, PPO evaluates this gradient through stochastic minibatch estimates, which we denote by $\mathbf{g}_{\mathrm{RL},\mathcal{B}} := \nabla_\phi \mathcal{L}_{\mathrm{PPO},\mathcal{B}}(\boldsymbol{\phi},\boldsymbol{\psi})$ for a minibatch $\mathcal{B}$.
At training iteration $k$, we denote by $\mathbf{g}_{\mathrm{RL},k}$ the realized stochastic PPO gradient used in that update, e.g., a minibatch estimate of the form $\mathbf{g}_{\mathrm{RL},\mathcal{B}_k}$.


\paragraph{Overview} We define the "Standard \gls{gbpl} Update" ($\mathcal{U}_\mathrm{GB}$, Eq.~\ref{eq:gb_update}) and "Standard RL Update" ($\mathcal{U}_\mathrm{RL}$, Eq.~\ref{eq:rl_update_std}) as the baselines. 
We propose four novel hybrid methods that inject the solver gradient $\mathbf{g}_\mathrm{SG}$ into different components of the \gls{ppo} architecture (Fig.~\ref{fig:bilevel_architecture}). We use one \gls{sgrl} variant per injection point so that the comparison isolates where solver guidance is most useful:
\begin{compactenum}
    \item \textbf{\gls{sgsca}:} Modulates the magnitude of the PPO policy update based on its alignment with a solver-consistent loss-gradient direction, acting as a dynamic learning rate.
    \item \textbf{\gls{sglos}:} Adds an auxiliary supervised loss that pulls the policy toward a solver-suggested target only when the RL agent performs poorly.
    \item \textbf{\gls{sgadv}:} Modifies the advantage function to reward action displacements that align with the negative solver loss gradient, biasing exploration.
    \item \textbf{\gls{sgcrt}:} Conditions the critic on the solver sensitivity vector, providing an additional landscape descriptor for value estimation.
\end{compactenum}
As shown in Fig.~\ref{fig:bilevel_architecture} and in Algorithms~\ref{alg:SGSCA}--\ref{alg:SGCRT}, each method represents a different strategy for fusing model-based \gls{nmpc} solver gradients with \gls{ppo}.

\subsection{Method 1: \glsentrylong{sgsca} (\glsentryshort{sgsca})}
Standard PPO already provides a loss-gradient update direction, but it still relies on a heuristic step size. We want the solver signal to tell us when PPO should trust that direction more strongly and when it should act more conservatively. We therefore use the \gls{nmpc} solver loss gradient to scale the magnitude of the PPO update while preserving the PPO update direction.

Because $\mathbf{g}_{\mathrm{SG},t}$ lives in the \gls{nmpc} weight space, we first define for each transition a \emph{solver-consistent scalar} whose gradient we can backpropagate through the policy:
\begin{equation}
    \label{eq:grad_pol_space}
    \ell_{\mathrm{SG},t}(\boldsymbol{\phi}) := \boldsymbol{\mu}_\phi(\boldsymbol{o}_t)^\top \mathbf{g}_{\mathrm{SG},t},
\end{equation}
During the PPO update, we treat $\mathbf{g}_{\mathrm{SG},t}$ as a fixed stored solver signal and differentiate only through the policy output $\boldsymbol{\mu}_\phi(\boldsymbol{o}_t)$. This induces the lifted parameter-space loss-gradient signal
\begin{equation}
\label{eq:grad_lifted_parameter_space}
    \widetilde{\mathbf{g}}_{\mathrm{SG},t} := \nabla_\phi \, \ell_{\mathrm{SG},t}(\boldsymbol{\phi}).
\end{equation}
At the PPO-objective level, we aggregate these per-transition quantities as
\begin{equation}
    \ell_\mathrm{SG}(\boldsymbol{\phi}) := \hat{\mathbb{E}}_t \left[ \ell_{\mathrm{SG},t}(\boldsymbol{\phi}) \right] = \hat{\mathbb{E}}_t \left[ \boldsymbol{\mu}_\phi(\boldsymbol{o}_t)^\top \mathbf{g}_{\mathrm{SG},t} \right],
\end{equation}
which induces the lifted loss-gradient signal
\begin{equation}
    \widetilde{\mathbf{g}}_\mathrm{SG} := \nabla_\phi \, \ell_\mathrm{SG}(\boldsymbol{\phi}).
\end{equation}
We then compute the alignment $\rho$ between the lifted solver-consistent loss-gradient signal and the PPO policy gradient $\mathbf{g}_\mathrm{RL}$ defined above:
\begin{equation}
    \rho = \frac{\widetilde{\mathbf{g}}_\mathrm{SG}^\top \mathbf{g}_\mathrm{RL}}{\|\mathbf{g}_\mathrm{RL}\|^2 + \epsilon}.
\end{equation}
This coefficient measures how strongly the solver-derived loss gradient projects onto the PPO loss gradient. We then construct an adaptive scaling factor $\alpha$:
\begin{equation} 
    \alpha = \text{clip}\left(1 + \lambda \cdot \rho, 0, \alpha_{\max}\right),
    \label{eq:adaptive_alpha}
\end{equation}
where $\lambda \geq 0$ is a strength hyperparameter and $\alpha_{\max}$ is a saturation limit preventing excessive update steps that could destabilize the policy. The lower bound of $0$ acts as a safeguard by preventing the adaptive mechanism from reversing the PPO update direction.
Finally, we scale the effective PPO update step and formally define the \glsentryshort{sgsca} operator $\mathcal{U}_{\mathrm{\glsentryshort{sgsca}}}$ as
\begin{equation}
    \mathcal{U}_{\mathrm{\glsentryshort{sgsca}}}: \boldsymbol{\phi}_{k+1} = \boldsymbol{\phi}_k - \eta \cdot \left( \colorbox{green!30}{$\alpha \cdot$} \mathbf{g}_\mathrm{RL} \right),
    \label{eq:sgsca_update}
\end{equation}

By scaling the magnitude of $\mathbf{g}_\mathrm{RL}$ in this way, we take larger descent steps $-\mathbf{g}_\mathrm{RL}$ when the solver loss gradient confirms the PPO loss gradient and smaller steps when it conflicts. In implementation, PPO evaluates both gradients through stochastic minibatch estimates on each minibatch $\mathcal{B}$, as summarized in Algorithm~\ref{alg:SGSCA} via $\ell_{\mathrm{SG},\mathcal{B}}$, $\widetilde{\mathbf{g}}_{\mathrm{SG},\mathcal{B}}$, and $\mathbf{g}_{\mathrm{RL},\mathcal{B}}$.

\begin{algorithm}[!t]
\caption{\glsentryshort{sgsca}}
\small
\begin{algorithmic}[1]
\For{each PPO minibatch}
    \State compute PPO minibatch gradient $\mathbf{g}_{\mathrm{RL},\mathcal{B}} = \nabla_\phi \mathcal{L}_{PPO,\mathcal{B}}$
    \State form $\ell_{\mathrm{SG},\mathcal{B}} = \sum_{i \in \mathcal{B}} \boldsymbol{\mu}_\phi(o_i)^\top \mathbf{g}_{SG,i}$ and $\widetilde{\mathbf{g}}_{\mathrm{SG},\mathcal{B}} = \nabla_\phi \ell_{\mathrm{SG},\mathcal{B}}$
    \State compute $\rho = (\widetilde{\mathbf{g}}_{\mathrm{SG},\mathcal{B}}^\top \; \mathbf{g}_{\mathrm{RL},\mathcal{B}})/(\|\mathbf{g}_{\mathrm{RL},\mathcal{B}}\|^2 + \epsilon)$ 
    \State compute $\alpha = \mathrm{clip}(1 + \lambda\rho, 0, \alpha_{\max})$
    \State update policy params $\boldsymbol{\phi}$ with scaled PPO gradient $\alpha\,\mathbf{g}_{\mathrm{RL},\mathcal{B}}$
\EndFor
\end{algorithmic}
\label{alg:SGSCA}
\end{algorithm}

\subsection{Method 2: \glsentrylong{sglos} (\glsentryshort{sglos})}
\gls{sglos} addresses the case in which PPO explores a poor weight configuration while the solver gradient suggests model-consistent local correction. At the same time, we do not want the solver signal to override trajectories that already achieve high return. We therefore add a performance-gated target-following loss to the minimized PPO training loss.
Unlike \gls{sgsca}, which rescales the PPO descent step, \gls{sglos} introduces an explicit \emph{target-following} loss $\mathcal{L}_{\mathrm{guide}}$ added to the PPO training loss:
\begin{equation} \label{eq:guided_loss}
    \mathcal{L}_{\mathrm{\gls{sglos}}}(\boldsymbol{\phi},\boldsymbol{\psi}) = \mathcal{L}_\mathrm{PPO}(\boldsymbol{\phi},\boldsymbol{\psi}) \colorbox{green!30}{$ + \lambda_{\mathrm{guide}} \, \mathcal{L}_{\mathrm{guide}}(\boldsymbol{\phi})$}.
\end{equation}
The surrogate loss $\tilde{\mathcal{L}}_{\mathrm{perf}}$ directly is not added to PPO as a globally active regularizer. Because that surrogate is local, open-loop, and model-based, applying it uniformly would treat the solver signal as equally reliable on all samples, including trajectories that already achieve high return. Instead, \gls{sglos} uses the solver gradient only to construct a local correction target and activates that correction selectively on underperforming transitions.
Let $\bar{\boldsymbol{\theta}}_t$ denote the behavior-policy weight vector associated with transition $t$. We form a local target weight configuration by taking a fixed step $\eta_{\mathrm{guide}}$ along the surrogate descent step $-\mathbf{g}_{\mathrm{SG},t}$:
\begin{equation}
    \boldsymbol{\theta}^{\mathrm{target}}_t = \bar{\boldsymbol{\theta}}_t - \eta_{\mathrm{guide}} \, \mathbf{g}_{\mathrm{SG},t}.
\end{equation}
We penalize the distance between the current policy mean $\boldsymbol{\mu}_\phi(\boldsymbol{o}_t)$ and this target with the gated loss
\begin{equation}
    \mathcal{L}_{\mathrm{guide}}(\boldsymbol{\phi}) = \hat{\mathbb{E}}_t \left[ w_t \, \| \boldsymbol{\mu}_\phi(\boldsymbol{o}_t) - \boldsymbol{\theta}^{\mathrm{target}}_t \|^2_2 \right].
\end{equation}
To implement this selective rectification, we derive the weight $w_t$ from the advantage function $\hat{A}_t$:
\begin{equation}
    w_t = \text{clip}(-\hat{A}_t, 0, w_{\max}),
\end{equation}
where $w_{\max}$ is the upper bound on the gating factor.
In implementation, we normalize the stored raw solver gradient, use the saved rollout mean as $\bar{\boldsymbol{\theta}}_t$ when it is available, and mask out transitions with negligible solver-gradient norm before forming the minibatch estimate of $\mathcal{L}_{\mathrm{guide}}$. With this gate, negative-advantage samples receive a corrective pull toward the solver-suggested target, whereas positive-advantage samples reduce to standard PPO. Early in training, when underperforming transitions are more common, the gate naturally activates more often; the overall aggressiveness is then controlled by $\lambda_{\mathrm{guide}}$, $\eta_{\mathrm{guide}}$, and the cap $w_{\max}$.

\begin{algorithm}[!t]
\caption{\glsentryshort{sglos}}
\small
\begin{algorithmic}[1]
\For{each collected transition}
    \State form $\boldsymbol{\theta}^{\mathrm{target}}_t = \bar{\boldsymbol{\theta}}_t - \eta_{\mathrm{guide}}\,\mathbf{g}_{\mathrm{SG},t}$ with the normalized solver gradient
    \State compute the gate $w_t = \text{clip}(-\hat{A}_t, 0, w_{\max})$
    \State ignore transitions with negligible solver-gradient norm
\EndFor
\For{each PPO minibatch}
    \State minibatch estimate of $\mathcal{L}_{\mathrm{guide}}$ using the valid transitions
    \State add $\lambda_{\mathrm{guide}}\,\mathcal{L}_{\mathrm{guide}}$ to the PPO training loss
    \State minimize the combined loss
\EndFor
\end{algorithmic}
\end{algorithm}

\subsection{Method 3: Advantage Shaping (\glsentryshort{sgadv})}
\gls{sgadv} targets exploration rather than the optimizer step. PPO learns through the advantage estimate $\hat{A}_t$, but $\hat{A}_t$ alone does not tell us whether the proposed policy-mean displacement points along a locally favorable solver descent step. We therefore inject the solver signal directly into the advantage estimate while keeping PPO on-policy.

As reflected in the PPO surrogate objective in Eq.~\eqref{eq:ppo_surrogate}, PPO optimizes the policy based on an estimate $\hat{A}_t$ of the advantage function. We augment this estimate with a gradient-based correction term derived from the alignment between the policy's \emph{mean weight displacement} and the local loss landscape in weight space.
We define the shaped advantage $\hat{A}^\text{SG}_t$ as:
\begin{equation}
    \hat{A}^\text{SG}_t = \hat{A}_t \colorbox{green!30}{$+ \beta \Delta_t^\mathrm{SG}$},
    \label{eq:sgadv_rule}
\end{equation}
where $\beta \geq 0$ is a scaling coefficient. The correction term $\Delta_t^\mathrm{SG}$ encodes the consistency between the policy's deterministic displacement from the rollout reference $\bar{\boldsymbol{\theta}}_t$ and the descent step $-\mathbf{g}_{\mathrm{SG},t}$:
\begin{equation}
    \Delta_t^\mathrm{SG} = - \left(\boldsymbol{\mu}_\phi(\boldsymbol{o}_t)-\bar{\boldsymbol{\theta}}_t\right)^\top \mathbf{g}_{\mathrm{SG},t}.
\end{equation}
Here, $\bar{\boldsymbol{\theta}}_t$ is the behavior-policy mean for the rollout transition. The term $\Delta_t^\mathrm{SG}$ represents the projection of the proposed displacement onto the descent step $-\mathbf{g}_{\mathrm{SG},t}$. Thus, $\Delta_t^\mathrm{SG} > 0$ implies that the policy mean moves locally toward lower surrogate performance loss, providing a solver-informed heuristic for action quality.
We then substitute the shaped advantage into the standard PPO surrogate optimization problem \eqref{eq:ppo_surrogate}:
\begin{equation}
\begin{aligned}
    \mathcal{J}^\mathrm{CLIP}_\mathrm{SG}(\boldsymbol{\phi}) = \hat{\mathbb{E}}_t \Big[ \min\big( &\rho_t(\boldsymbol{\phi}) \hat{A}^\text{SG}_t, \\
    &\text{clip}(\rho_t(\boldsymbol{\phi}), 1-\epsilon, 1+\epsilon) \hat{A}^\text{SG}_t \big) \Big].
\end{aligned}
\end{equation}
With this shaping, we reward solver-consistent action displacements and penalize solver-inconsistent ones through the advantage signal itself.

\begin{algorithm}[!t]
\caption{\glsentryshort{sgadv}}
\small
\begin{algorithmic}[1]
\For{each collected transition}
    \State compute $\Delta_t^\mathrm{SG} = -\left(\boldsymbol{\mu}_\phi(\boldsymbol{o}_t)-\bar{\boldsymbol{\theta}}_t\right)^\top \mathbf{g}_{\mathrm{SG},t}$
    \State set $\hat{A}^\mathrm{SG}_t = \hat{A}_t + \beta\,\Delta_t^\mathrm{SG}$
\EndFor
\For{each PPO minibatch}
    \State replace $\hat{A}_t$ by $\hat{A}^\mathrm{SG}_t$ in the clipped PPO surrogate
    \State run the standard PPO actor--critic update with $\hat{A}^\mathrm{SG}_t$
\EndFor
\end{algorithmic}
\end{algorithm}


\subsection{Method 4: \glsentrylong{sgcrt} (\glsentryshort{sgcrt})}
\gls{sgcrt} targets the critic rather than the actor. The same observation can correspond to very different \emph{local controllability and tuning difficulty} depending on how sensitive the \gls{nmpc} solution is to the cost weights, so an observation-only critic misses useful information about the local landscape. We therefore augment the critic with a solver-derived \emph{landscape descriptor} based on the weight-space loss gradient $\mathbf{g}_\mathrm{SG}$ and inject it through a residual correction branch:

\begin{equation}
    V_\psi\big(\boldsymbol{o}_t, \mathbf{g}_{\mathrm{SG},t^-}\big)
    = V_{\boldsymbol{\psi},\mathrm{base}}(\boldsymbol{o}_t)
    \; \colorbox{green!30}{$+ \; c_{\psi}\!\left(\mathbf{g}_{\mathrm{SG},t^-}\right)$},
\end{equation}
where $V_{\boldsymbol{\psi},\mathrm{base}}$ is the conventional critic (MLP neural network), and $c_{\psi}$ is a lightweight gradient encoder, e.g., a small fully connected network with bounded scalar output, that maps the normalized solver gradient to a residual correction.
We use the index $t^-$ to emphasize that, in an online loop, the critic uses the \emph{most recently available} solver sensitivity (e.g., from the previous \gls{nmpc} solve) when evaluating the current observation. For the policy-gradient update, $\mathbf{g}_{\mathrm{SG},t^-}$ is treated as a detached, action-independent critic feature, and no actor gradient is propagated through the value target or residual branch.
\begin{figure}[!t]
    \centering
    \includegraphics[width=0.9\columnwidth]{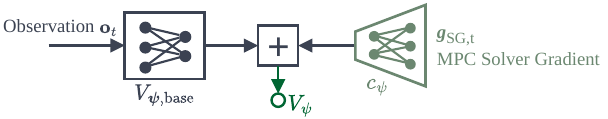}
    \caption{Residual \glsentryshort{sgcrt} critic: an observation-based value estimate plus a scalar correction from a solver-gradient encoder.}
    \label{fig:critic_architecture}
\end{figure}
With this residual branch, we calibrate value predictions in high-sensitivity regions and can reduce the advantage variance without modifying PPO's actor update rule.

\begin{algorithm}[!t]
\caption{\glsentryshort{sgcrt}}
\small
\begin{algorithmic}[1]
\For{each environment step}
    \State cache the latest available solver gradient $\mathbf{g}_{\mathrm{SG},t^-}$
    \State evaluate $V_\psi(\boldsymbol{o}_t, \mathbf{g}_{\mathrm{SG},t^-}) = V_{\boldsymbol{\psi},\mathrm{base}}(\boldsymbol{o}_t) + c_\psi(\mathbf{g}_{\mathrm{SG},t^-})$
\EndFor
\For{each PPO minibatch}
    \State compute returns and advantages with the augmented critic
    \State actor update with standard PPO and augmented critic fitting
\EndFor
\end{algorithmic}
\label{alg:SGCRT}
\end{algorithm}

\paragraph*{Weight-space vs. parameter-space usage}
The solver signal $\mathbf{g}_{\mathrm{SG},t}$ is an \gls{nmpc} \emph{weight-space} loss gradient. \gls{sglos}, \gls{sgadv}, and \gls{sgcrt} use it directly in weight space (to shape advantages, construct targets, or condition the critic). \gls{sgsca}, in contrast, needs a policy \emph{parameter-space} quantity to modulate the PPO update step: it therefore lifts the weight-space signal through the policy using the construction already introduced in Eq.~\eqref{eq:grad_pol_space}.
\section{Optimization Interpretation of \gls{sgrl}}
\label{sec:theory}

\begin{figure*}[!t]
\centering
\includegraphics[width=0.80\textwidth]{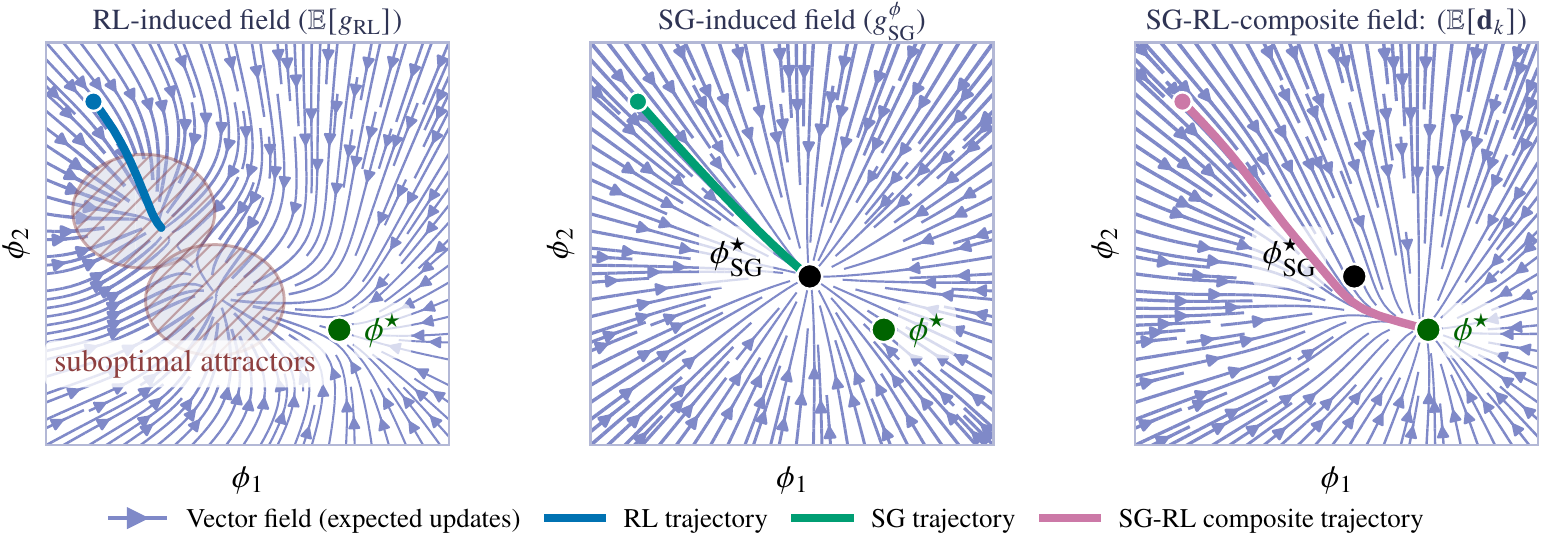}
\caption{Idealized policy-parameter dynamics. RL gradients optimize closed-loop return but may get trapped in poor basins; solver gradients provide biased local guidance; SG-RL combines both to guide early learning before RL refinements dominate.}
\label{fig:sgrl_geometry_flow}
\end{figure*}

We formalize the \gls{sgrl} update in the same descent-direction convention used above and analyze its relationship to the expected closed-loop return $\mathcal{R}(\pi_{\boldsymbol{\phi}})$. Since the reward in Section~\ref{sec:rl_approach} is defined as $r_t=-\mathcal{L}_{\mathrm{perf}}(\mathbf{s}_t,\mathbf{u}_0^\star)$, maximizing $\mathcal{R}(\pi_{\boldsymbol{\phi}})$ is equivalent to minimizing the expected cumulative loss in Eq.~\eqref{eq:closed_loop_learning_objective}. We interpret $\mathbf{g}_{\mathrm{RL},k}$ as the iteration-$k$ stochastic realization of the PPO loss gradient in Eq.~\eqref{eq:g_RL} and treat it as a stochastic descent proxy for $-\nabla_{\boldsymbol{\phi}}\mathcal{R}(\pi_{\boldsymbol{\phi}_k})$.

This section provides a first-order optimization interpretation of \gls{sgrl} rather than formal convergence guarantees. For analytical tractability, we discuss \gls{sgrl} through descent directions, while the implementation optimizes PPO's clipped surrogate as a trust-region approximation. The generic update model below provides a common interpretive lens for the \gls{sgrl} family; the individual methods differ only in how solver-gradient information is incorporated into the PPO optimization.

\subsection{Solver Guidance Signal and \gls{sgrl} Update}
Using the same lifting construction as in Eqs.~\eqref{eq:grad_pol_space} and~\eqref{eq:grad_lifted_parameter_space}, we denote the model-based \gls{nmpc} solver gradient in weight space at a generic observation $\boldsymbol{o}$ by
\begin{equation}
    \mathbf{g}_\mathrm{SG}^\theta(\boldsymbol{o}) :=
    \nabla_{\boldsymbol{\theta}}
    \tilde{\mathcal{L}}_\text{perf}
    \bigl(\mathbf{z}^*(\boldsymbol{\mu}_\phi(\boldsymbol{o}))\bigr).
\end{equation}
We denote by $\mathbf{g}_\mathrm{SG}^\phi(\boldsymbol{o})$ the corresponding lifted normalized loss-gradient in policy-parameter space. The descent direction is therefore $-\mathbf{g}_\mathrm{SG}^\phi$, which should be interpreted as a local model-based guidance signal induced by the surrogate $\tilde{\mathcal{L}}_\text{perf}$ rather than as the true policy gradient of $\mathcal{R}(\pi_{\boldsymbol{\phi}})$.

We define the \gls{sgrl} update as $\boldsymbol{\phi}_{k+1} = \boldsymbol{\phi}_k - \alpha_k \mathbf{d}_k$, where the composite loss-gradient update vector $\mathbf{d}_k$ is given by:
\begin{equation} \label{eq:composite_direction}
\mathbf{d}_k = \mathcal{F}_k(\mathbf{g}_{\mathrm{RL},k},\,\mathbf{g}_{\mathrm{SG},k}^\phi).
\end{equation}
The operator $\mathcal{F}_k$ combines the PPO and solver-guidance signals into a single update direction (Fig.~\ref{fig:sgrl_teaser}). The iteration index permits dynamic blending strategies, such as annealing solver influence through
$
    \mathcal{F}_k(\mathbf{x},\mathbf{y}) = \mathbf{x} + \lambda w_k \mathbf{y},
    \quad
    w_k\in[0,1].
$

\subsection{Early-Training Guidance}
\label{sec:speedup}
The solver-guidance signal $\mathbf{g}_{\mathrm{SG},k}^\phi$ is obtained by differentiating the surrogate $\tilde{\mathcal{L}}_\mathrm{perf}$ rather than the expected return, and therefore introduces model bias. Conversely, $\mathbf{g}_{\mathrm{RL},k}$ is estimated from sampled closed-loop trajectories and typically exhibits much higher trajectory-level variance early in training. \gls{sgrl} exploits this complementary bias--variance trade-off by using solver gradients only as guidance toward promising regions of the policy landscape (Fig.~\ref{fig:sgrl_geometry_flow}).

To make this intuition explicit, let
\begin{equation}
    \boldsymbol{\phi}^\star_\mathrm{SG}
    \in
    \argmin_{\boldsymbol{\phi}}\,
    \hat{\mathbb{E}}_{\boldsymbol{o}\sim \mathcal{D}} \left[
    \tilde{\mathcal{L}}_\mathrm{perf}\bigl(\mathbf{z}^*(\boldsymbol{\mu}_\phi(\boldsymbol{o}))\bigr)
    \right]
\end{equation}
be surrogate-optimal policy parameters under a generic dataset $\mathcal{D}$ of observations collected during training. The surrogate is useful whenever optimizing it yields policies that outperform random initialization under the true return, i.e.,
\begin{equation}
    \mathcal{R}(\pi_{\boldsymbol{\phi}^\star_\mathrm{SG}}) > \mathbb{E}_{\boldsymbol{\phi}_0 \sim \mathcal{P}_0}\left[\mathcal{R}(\pi_{\boldsymbol{\phi}_0})\right].
\end{equation}
Here, $\boldsymbol{\phi}_0 \sim \mathcal{P}_0$ denotes a randomly initialized policy parameter set.
The surrogate need not share the true optimum. It only needs to provide informative guidance during early learning. Our \gls{sgsca}, \gls{sglos}, and \gls{sgadv} methods use this surrogate bias in actor-side training signals. In contrast, \gls{sgcrt} uses solver sensitivities only as additional critic features. When detached from actor-gradient paths, the additional critic features leave the expected policy-gradient estimator unchanged while potentially reducing variance through a more accurate value baseline.

Because the actor-side methods intentionally bias optimization toward the surrogate objective, their influence is gradually annealed by scheduling $(\lambda,\lambda_{\mathrm{guide}},\beta)$ in Eq.~\eqref{eq:adaptive_alpha}, Eq.~\eqref{eq:guided_loss}, and Eq.~\eqref{eq:sgadv_rule}. This heuristic progressively transfers control back to PPO as learning progresses.

\subsection{Non-aligned Updates}
\label{sec:return_aligned_updates}

During this transition, the composite update $\mathbf{d}_k$ generally differs from the pure PPO direction. It therefore need not maximize immediate return improvement, but it can accelerate optimization by steering learning toward more promising regions before PPO dominates (Fig.~\ref{fig:sgrl_geometry_flow}).
To make it concrete, consider \gls{sgsca} in a near-plateau regime where $\mathbb{E}[\mathbf{g}_{\mathrm{RL},k}] \approx \mathbf{0}$. Because the scaling factor $\alpha_k$ in \eqref{eq:adaptive_alpha} depends directly on the minibatch gradient $\mathbf{g}_{\mathrm{RL},k}$, the expectation of their product does not factorize:
\begin{equation}
    \mathbb{E}[\alpha_k \, \mathbf{g}_{\mathrm{RL},k}] = \mathbb{E}[\alpha_k] \, \mathbb{E}[\mathbf{g}_{\mathrm{RL},k}] + \mathrm{Cov}(\alpha_k, \mathbf{g}_{\mathrm{RL},k}).
\end{equation}
Near such a plateau, the covariance term can induce a nonzero expected drift by amplifying solver-aligned minibatch directions and suppressing conflicting ones. Still, the closed-loop return objective $\mathcal{R}(\pi_{\boldsymbol{\phi}})$ remains the primary optimization signal.

\begin{figure*}[!t]
\centering
\includegraphics[width=\textwidth]{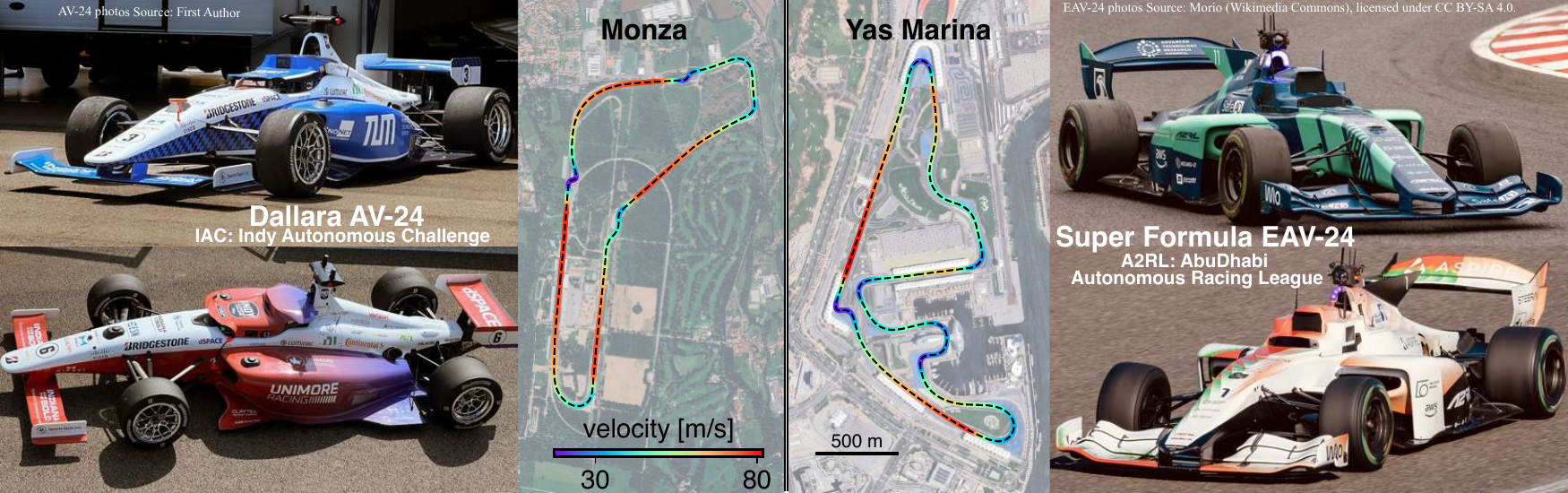}
\caption{Autonomous-racing validation platforms: Dallara AV-24 on Monza (left) and Super Formula EAV-24 on Yas Marina (right), both reaching peak velocities above \SI{80}{\meter\per\second}.}
\label{fig:multi_track_velocity_heatmap}
\end{figure*}
\section{Application Example: High-Speed Autonomous Racing}
\label{sec:application_AV}
We validate \gls{sgrl} in high-speed autonomous racing on two full-scale, high-fidelity simulation platforms representative of contemporary competitions: the Dallara AV-24 used in the Indy Autonomous Challenge (IAC) and the Super Formula EAV-24 from the Abu Dhabi Autonomous Racing League (A2RL). Both platforms are based on simulation and \gls{nmpc} models identified from real-world vehicle data, while their underlying \gls{nmpc} controllers have already been validated on the physical race cars.

\subsection{Why Autonomous Racing Stresses Weights Tuning}
Autonomous racing is a demanding benchmark for weights-varying \gls{nmpc} because it combines tight feasibility constraints at the actuation and tire limits, pronounced modeling uncertainty and nonlinearity, and strong regime shifts within a single lap. Indeed, racing combines qualitatively different regimes: straights demand accurate velocity tracking, braking zones require deceleration stability, corners require lateral accuracy under combined-acceleration limits, and corner exits prioritize traction-limited acceleration. A single static weight tuning must compromise across these regimes, motivating online adaptation of the weight vector $\boldsymbol{\theta}_t$.

\subsection{Vehicle Dynamic Model for \gls{nmpc}}
In our \gls{nmpc} problem \eqref{eq:nmpc_ocp}, we employ a nonlinear single-track vehicle model $\dot{\mathbf{x}} = \boldsymbol{f}(\mathbf{x}, \mathbf{u})$ based on \cite{zarrouki2023,zarrouki_rnmpc_2024}. The states $\mathbf{x} = [x, y, \psi, v_x, v_y, \dot{\psi}, \delta, a_x]^\top$ model the vehicle pose, velocities, steering angle, and longitudinal acceleration, while the inputs $\mathbf{u} = [j_x, \dot{\delta}]^\top$ are the longitudinal jerk and steering rate. Lateral tire forces are modeled by the nonlinear Pacejka Magic Formula \cite{pacejka1997magic}, and aerodynamic drag and downforce are included explicitly, affecting both longitudinal acceleration limits and available tire grip.

\subsection{\gls{nmpc} Formulation}

The \gls{nmpc} tracks a time-optimal trajectory (path and velocity profile) generated by an optimal control planner \cite{rowold_2023}. At each control step, the planner provides an $N$-step reference segment containing desired position, heading, velocity, longitudinal acceleration, and lateral acceleration. The \gls{nmpc} problem follows the structure in Section~\ref{sec:problem_formulation}. 
The cost function consists of a stage cost $l(\cdot)$ and a terminal cost $l_N(\cdot)$, formulated as quadratic forms:
\begin{align}
    l(\mathbf{x}_k, \mathbf{u}_k, \boldsymbol{\theta}_t) &= \frac{1}{2} \, \mathbf{r}(\mathbf{x}_k, \mathbf{u}_k)^\top \mathbf{W}(\boldsymbol{\theta}_t) \, \mathbf{r}(\mathbf{x}_k, \mathbf{u}_k), \\
    l_N(\mathbf{x}_N, \boldsymbol{\theta}_t) &= \frac{1}{2} \,\mathbf{r}_N(\mathbf{x}_N)^\top \mathbf{W}_N(\boldsymbol{\theta}_t) \, \mathbf{r}_N(\mathbf{x}_N).
\end{align}
Here, $\mathbf{r}(\cdot)$ denotes the vector of tracking errors and regularization terms. The weighting matrix $\mathbf{W}(\boldsymbol{\theta}_t) = \text{diag}(\boldsymbol{\theta}_t)$ is directly parameterized by the policy action vector $\boldsymbol{\theta}_t$, allowing our cost-weight policy to reshape the quadratic cost landscape online. The weights are fixed along the prediction horizon but updated between \gls{nmpc} solves.

The feature vector for the stage cost is defined as:
\begin{equation}
    \mathbf{r}(\mathbf{x}, \mathbf{u}) = \begin{bmatrix} e_{\mathrm{lat}}, & e_{\psi}, & e_{v}, & e_{a}, &  e_{a_{\mathrm{lat}}}, & u_{j}, & u_{\dot{\delta}} \end{bmatrix}^\top.
\end{equation}
Here, $e_{\mathrm{lat}}$, $e_{\psi}$, $e_{v}$, $e_{a}$, and $e_{a_{\mathrm{lat}}}$ denote the tracking errors for lateral position, heading, velocity, longitudinal acceleration, and lateral acceleration, while $u_j$ and $u_{\dot{\delta}}$ penalize high-frequency actuation. The terminal feature vector $\mathbf{r}_N(\mathbf{x}_N)$ contains the same state-dependent terms but excludes the control inputs.

The learnable weight vector (policy action) is
\begin{equation}
    \label{eq_weights_NMPC}
	\boldsymbol{\theta}_t := \begin{bmatrix}
		q_{\ell,t} & q_{\psi,t} & q_{v,t} & q_{a,t} & q_{a_y,t} & r_{j,t} & r_{\dot{\delta},t}
	\end{bmatrix}^\top \in \mathbb{R}^{7}_{>0},
\end{equation}
representing the weights on the corresponding error terms in the cost function. 
We enforce strict feasibility bounds on actuators and physical envelopes on the states:

\paragraph{Actuation Limits} The control inputs are bounded to respect the physical steering rate and jerk limits:
\begin{equation}
    \mathcal{U} = \{ \mathbf{u} \mid \dot{\delta}_{\min} \le \dot{\delta} \le \dot{\delta}_{\max}, \quad j_{\min} \le j_x \le j_{\max} \}.
\end{equation}

\paragraph{State Constraints} The steering angle $\delta$ is limited by the mechanical steering range. The longitudinal acceleration $a_x$ is restricted by the velocity-dependent powertrain envelope, capturing the engine power curve and braking capacity:    \\ $\delta_{\min} \le \delta_k \le \delta_{\max}$ and $a_{x,\min}(v_{x,k}) \le a_{x,k} \le a_{x,\max}(v_{x,k})$.

\paragraph{Combined Friction Ellipse (GG-Diagram)} We impose a coupling constraint between longitudinal acceleration $a_x$ and lateral acceleration $a_y \approx v_x \dot{\psi}$.
We approximate the friction utilization using a generalized ellipsoid \cite{Piccinini2024_ggv}: $\left( | a_{x,k} |/a_{x,\max}^{\mathrm{comb}}(v_{x,k}) \right)^{\eta} + \left( | a_{y,k} |/a_{y,\max}^{\mathrm{comb}}(v_{x,k}) \right)^{\eta} \le 1$.
Here, $\eta$ shapes the envelope and $a_{x/y,\max}^{\mathrm{comb}}(v_{x,k})$ are velocity-dependent acceleration capacities.
\subsection{Observations for Racing}

The policy input $o_t$ must support reliable weight selection while staying low-dimensional and transferable. We use three information sources:
    \paragraph{Current motion state (stacked measurements)} We include a stack of the most recent longitudinal velocities $\big\{ v_x(t-i) \big\}_{i=0}^{n_h-1}$ and yaw rates $\big\{ r(t-i) \big\}_{i=0}^{n_h-1}$.
    Here, $r$ denotes yaw rate and $n_h$ is the stack length. This captures the current dynamic regime and short transients.
    \paragraph{Performance history (tracking deviations)} We include a stack of recent magnitudes of path-tracking deviations $\big\{ |e_{\mathrm{lat}}(t-i)| \big\}_{i=0}^{n_h-1}$ and velocity-tracking deviations $\big\{ |e_v(t-i)| \big\}_{i=0}^{n_h-1}$.
   \paragraph{Task anticipation (look-ahead)} We include samples of the reference velocity $\big\{ v_{\mathrm{ref}}(t+\tau_j) \big\}_{j=1}^{K}$ and path curvature $\big\{ \kappa_{\mathrm{ref}}(t+\tau_j) \big\}_{j=1}^{K}$,
    with $K$ anticipation points over the \gls{nmpc} horizon, where $\tau_j$ are uniformly spaced look-ahead times in $[0,T_p]$. 


\subsection{Learning Cost-Weight Policies}
We treat the \gls{nmpc} closed loop as the environment and bounded weights $\boldsymbol{\theta}_t$ as actions. At each decision time, the actor outputs $\boldsymbol{\theta}_t \in [\boldsymbol{\theta}_{\min},\boldsymbol{\theta}_{\max}]$, parameterizing $\mathbf{W}(\boldsymbol{\theta}_t)$ and $\mathbf{W}_N(\boldsymbol{\theta}_t)$.
%
The differentiable \gls{nmpc} solver provides primal-solution sensitivities with respect to the weights, enabling the closed-loop solver-derived gradient $\mathbf{g}_{SG,t}$ used by the \gls{sgrl} mechanisms. 
We evaluate each closed-loop transition using a fixed per-step performance loss $\mathcal{L}_{\mathrm{perf}}(\mathbf{s}_t, \mathbf{u}^*_0)$:
\begin{equation}
    \begin{split}
        \mathcal{L}_{\mathrm{perf}}(\mathbf{s}_t, \mathbf{u}^*_0) = {}& w_v e_{v,t}^2 + w_{\mathrm{lat}} e_{\mathrm{lat},t}^2 \\
        & {}+ w_j \mathcal{P}_{\exp}(j_t, \bar{j}) + w_{\dot{\delta}} \mathcal{P}_{\exp}(\dot{\delta}_t, \bar{\dot{\delta}}).
    \end{split}
    \label{eq:L_perf}
\end{equation}
Here, $e_{v,t}$ and $e_{\mathrm{lat},t}$ are realized velocity/lateral errors, and $j_t$, $\dot{\delta}_t$ are realized jerk and steering rate. The function $\mathcal{P}_{\exp}(u, \bar{u}) = \exp(\lambda_{\mathrm{exp}}(|u| - \bar{u})) - 1$ if $|u| > \bar{u}$ and $u^2$ otherwise, with $\lambda_{\mathrm{exp}}>0$, strongly penalizes violations of $\bar{j}$ and $\bar{\dot{\delta}}$. The weights $\mathbf{G}_{\mathrm{perf}} = \{w_v, w_{\mathrm{lat}}, w_j, w_{\dot{\delta}}\}$ are fixed, and $r_t = -\mathcal{L}_{\mathrm{perf}}(\mathbf{s}_t, \mathbf{u}^*_0)$ aligns the RL reward with the solver-gradient signal.

\section{Results}
\begin{figure*}[!t]
    \centering
    \includegraphics[width=\textwidth]{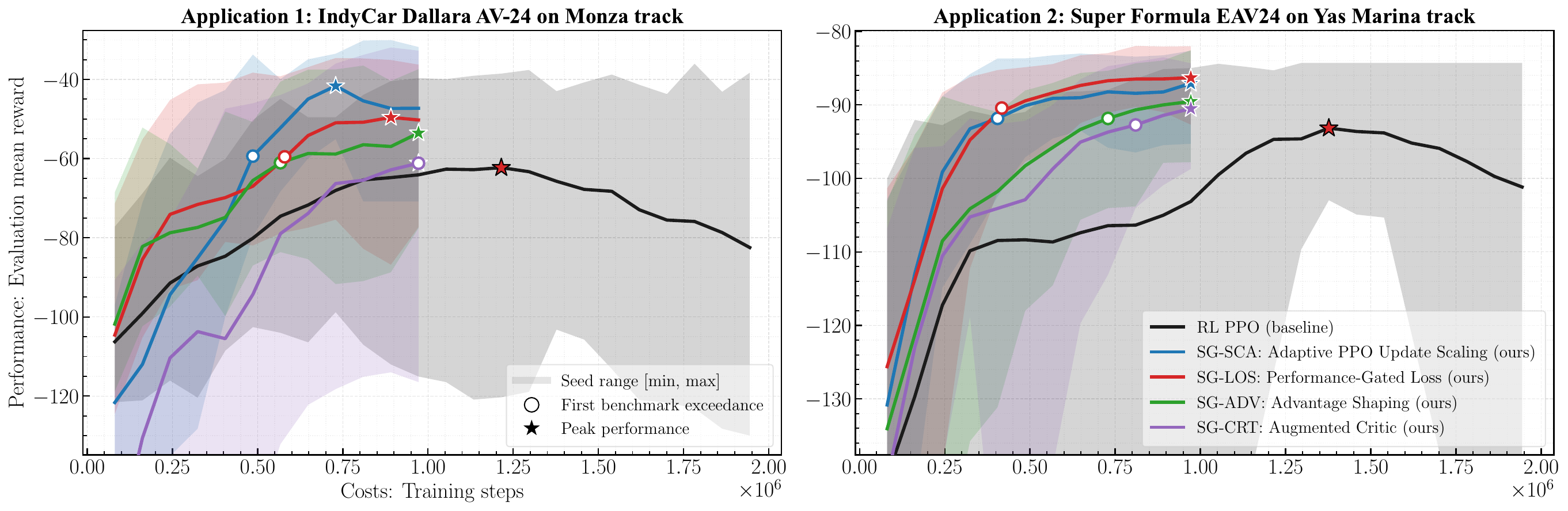}
    \caption{Deterministic evaluation return during training over 10 seeds. Lines show seed means, shaded bands show seed ranges, stars mark best mean returns, and circles mark when \gls{sgrl} first surpasses PPO's best mean return.}
    \label{fig:sg_sweeps_eval_rewards_combined}
\end{figure*}
\subsection{Experimental Setup}
We train and evaluate cost-weight policies in closed-loop simulation on the AV-24 Indy Dallara and EAV-24 Super Formula platforms. Both high-fidelity simulation and \gls{nmpc} prediction models were identified from real vehicle-dynamics data and validated against racing-event measurements~\cite{Sagmeister2024,Musiu22012026}.
The exact \gls{nmpc} implementations used in this work were transferred to the real cars and validated during full-scale racing at Yas Marina and Modena in April/June 2025 with the same Weights-varying architecture and a rule-based policy\footnote{Supplementary real-world evidence:
\url{anonymous.4open.science/r/WMPC}
}. Since only the weight-selection policy changes, learned policies are directly transferable to the validated controller stack. Learning remains in simulation because near-limit exploration on full-scale race cars is unsafe and costly.

A higher-level planner provides the local raceline, and the differentiable \gls{nmpc} computes inputs for the high-fidelity plant simulator. Plant and prediction models are nonlinear single-track models with nonlinear tires but intentionally differ in fidelity (Table~\ref{tab:plant_vs_predictor}), creating a deployment-relevant \textbf{model-mismatch setting}.
Platform-specific parameters and actuator bounds are listed in Table~\ref{tab:platform_vehicle_params}.
\begin{table}[!t]
\centering
\caption{Key differences between the nonlinear single-track simulation and \gls{nmpc} prediction models.}
\label{tab:plant_vs_predictor}
\small
\begin{tabular}{lcc}
\hline
 & \textbf{Sim. Model} & \textbf{\gls{nmpc} Pred. Model} \\
\hline
Inputs & \shortstack{$a_x$, $\delta$} & \shortstack{$j_x$, $\dot{\delta}$} \\
\hline
Actuator & \shortstack{Explicit with} & \shortstack{simplified} \\
dynamics & \shortstack{rate saturation} &  \\
\hline
 Tire & \shortstack{Wheel-level,} & \shortstack{Axle-level,} \\
 model& \shortstack{Pacejka MF96 \cite{pacejka1997magic}} & \shortstack{reduced Pacejka MF \cite{pacejka1997magic}} \\
 & combined slip & \shortstack{no combined-slip} \\
\hline
 Load & \shortstack{Full, with} & \shortstack{Simplified, mainly} \\
 transfer & \shortstack{roll, lateral and} & \shortstack{longitudinal} \\
 & \shortstack{longitudinal effects} &  \\
\hline
 Aerodyn. & \shortstack{Full drag,} & \shortstack{Simplified drag,} \\
 & \shortstack{front/rear downforce,} & \shortstack{lumped downforce,} \\
 & \shortstack{velocity-dependent} & \shortstack{no vel. dependency} \\
\hline
\end{tabular}
\end{table}
\begin{table}[!t]
\centering
\caption{Single-track vehicle parameters and actuation bounds for the Dallara AV-24 and EAV-24 Super Formula.}
\label{tab:platform_vehicle_params}
\small
\begin{tabular}{lcc}
\hline
\textbf{Parameter} & \textbf{AV-24} & \textbf{EAV-24} \\
\hline
Wheelbase $l$ [m] & 2.971 & 3.115 \\
CG--front axle $l_f$ [m] & 1.724 & 1.802 \\
Mass $m$ [kg] & 800 & 782 \\
Yaw inertia $I_z$ [kg\,m$^2$] & 1000 & 700 \\
Front track $b_f$ [m] & 1.639 & 1.606 \\
Rear track $b_r$ [m] & 1.524 & 1.510 \\
Steering $\delta$ [rad] & $[-0.276,\,0.276]$ & $[-0.342,\,0.342]$ \\
Steering rate $\dot\delta$ [rad/s] & $[-0.429,\,0.429]$ & $[-1.050,\,1.050]$ \\
\hline
\end{tabular}
\end{table}
The platforms differ in aerodynamics, load transfer, actuation envelopes, tires, and mass/inertia, making the weight-to-behavior map platform-dependent and requiring distinct bounds and policies.
We obtained the bounds for the 7-dimensional weight vector $\boldsymbol{\theta}_t$ in \eqref{eq_weights_NMPC} empirically from platform-specific fixed-weight \gls{nmpc} tuning. 
\paragraph{Benchmarks} We benchmark our four proposed \gls{sgrl} methods against four reference baselines: the \gls{rl} PPO baseline, a fixed-weight human-expert \gls{nmpc} controller, the reduced \gls{gbpl} baseline from \cite{2025GB_PL_DiffNMPC}, and our \gls{gbpl} baseline.

\paragraph{\gls{nmpc} solver configuration}
We solve the \gls{nmpc} problem with \texttt{acados} \cite{verschueren_acadosmodular_2022} using a real-time SQP-RTI scheme and full-condensing HPIPM as the QP backend. QP iterations were limited to 25 and equality and stationarity tolerances were set to $10^{-3}$.

\paragraph{Timing, horizons, and rollout length}
The plant simulator runs at $T_{\mathrm{s,sim}}=\SI{0.02}{s}$.
The \gls{nmpc} optimal control problem is discretized with $T_{\mathrm{s}}=\SI{0.075}{s}$ over a prediction horizon of $T_{\mathrm{p}}=\SI{2.6}{s}$, yielding $N=\lfloor T_{\mathrm{p}}/T_{\mathrm{s}}\rfloor=34$ shooting intervals.
At each simulator step (\(T_{\mathrm{s,sim}}=\SI{0.02}{\second}\)), the policy outputs weights, \gls{nmpc} solves, and the first input is applied. Each \(T_{\mathrm{roll}}=\SI{135}{\second}\) rollout therefore has \(6750\) simulator steps/policy decisions/control updates/solves, while prediction discretization remains \(T_s=\SI{0.075}{\second}\). Episodes terminate after repeated solver failures or track departures, with a training-only truncation penalty.
\paragraph{Learning algorithm and hyperparameters}
\label{paragraph:hyperparams}

For fairness, PPO and all \gls{sgrl} methods share the same \gls{rl} setup. Only \gls{gbpl} differs. \gls{rl}-based methods use Stable-Baselines3 PPO~\cite{stable-baselines3} with a Gaussian MLP actor--critic, bounded normalized outputs de-normalized to platform-specific weight bounds, generalized state-dependent exploration, 12 parallel environments, and deterministic evaluation every rollout. PPO trains for 2M steps, whereas \gls{sgrl} trains for 1M to test whether solver guidance halves the sample budget. Because \gls{gbpl} becomes unstable with large rollouts/batches, we follow~\cite{2025GB_PL_DiffNMPC} and use 10-transition fragments/minibatches with one training environment. The \gls{rl} reward is $-\mathcal{L}_{\mathrm{perf}}$ (Eq.~\eqref{eq:L_perf}).
\paragraph{Policy initialization}
We initialize policy linear layers with orthogonal weights and zero biases, a standard PPO choice that balances initial signal scales. Unlike~\cite{2025GB_PL_DiffNMPC}, which uses a pre-optimized output bias, we set this bias to zero so the initial normalized mean action lies near the admissible hyper-rectangle center.
\paragraph{Hardware}
Experiments ran on a 2022 MacBook Air (Apple M2, \SI{16}{GB} RAM): 
PPO took \SI{45}{min} for 2M training samples and \gls{sgrl} \SI{20}{min} for 1M; on Apple M4, the times dropped to \SI{28}{min} and \SI{14}{min}.
\subsection{Key Highlights: Performance and Sample Efficiency}
\begin{figure}[!t]
    \centering
    \includegraphics[width=0.95\columnwidth]{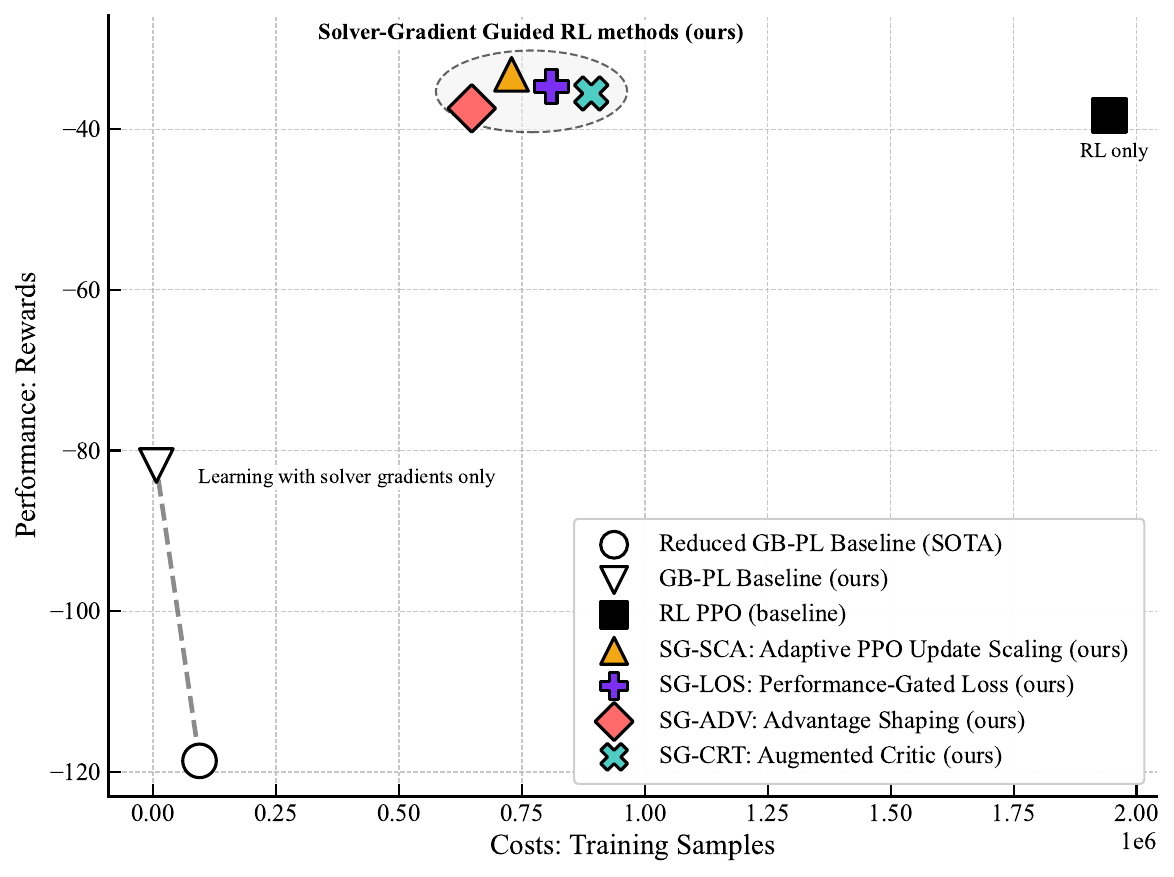}
        \caption{Best deterministic evaluation return versus training samples for the AV-24/Monza budget comparison. 
        }
    \label{fig:results_beat_baseline_vs_best}
\end{figure}
Figs.~\ref{fig:sg_sweeps_eval_rewards_combined} and~\ref{fig:results_beat_baseline_vs_best} summarize closed-loop performance versus sample efficiency. On AV-24/Monza, PPO reaches its highest mean return ($-62.29$) after $1.215\cdot 10^6$ samples, whereas all four \gls{sgrl} methods surpass it earlier: \gls{sgsca} after $4.86\cdot 10^5$ samples ($60.0\%$ fewer), \gls{sglos}/\gls{sgadv} after $5.67\cdot 10^5$ ($53.3\%$ fewer), and \gls{sgcrt} after $6.48\cdot 10^5$ ($46.7\%$ fewer). \gls{sgsca} also improves PPO's peak mean reward by $33.2\%$, while \gls{sglos}, \gls{sgcrt}, and \gls{sgadv} gain $20.4\%$, $18.5\%$, and $14.1\%$.

The same picture holds on EAV-24/Yas Marina, with even larger relative sample-efficiency gains for the strongest methods. \gls{sgsca} and \gls{sglos} both surpass PPO's highest mean return after only $4.05\cdot 10^5$ samples, i.e., $70.6\%$ earlier, whereas \gls{sgadv} and \gls{sgcrt} still retain $47.1\%$ and $35.3\%$ sample savings. The min--max bands also suggest improved training stability under solver-gradient guidance: PPO shows the widest persistent scatter in reward evolution, whereas \gls{sgsca} and \gls{sglos} concentrate earlier around high-return regimes. Taken together, the mean training curves show that solver-gradient guidance consistently improves PPO sample efficiency, with \gls{sgsca} and \gls{sglos} providing the strongest cross-platform trade-off.

Fig.~\ref{fig:results_beat_baseline_vs_best} complements this cross-seed view with a single-seed AV-24/Monza budget comparison. All \gls{sgrl} methods remain far above our \gls{gbpl} baseline, with best-return gains of $54.3\%$--$59.5\%$ over our \gls{gbpl} and $68.5\%$--$72.0\%$ over the reduced \gls{gbpl} baseline. At the same time, they require only $8.10\cdot 10^5$ to $9.72\cdot 10^5$ samples, i.e., $50.0\%$--$58.3\%$ fewer than PPO's $1.94\cdot 10^6$ budget. \gls{sgsca} and \gls{sglos} provide the clearest trade-off, combining the lowest \gls{sgrl} sample counts with the strongest returns relative to \gls{gbpl}. These results support the grey-box thesis: solver sensitivities improve exploration efficiency without sacrificing PPO stability.

\label{sec:results}
Within the \gls{gbpl} family, Table~\ref{tab:performance_overview} shows that our \gls{gbpl} baseline also substantially outperforms the reduced \gls{gbpl} baseline from \cite{2025GB_PL_DiffNMPC}: total return improves from $-118.62$ to $-81.87$, i.e., by 31.0\%, while the required training data drop from 94K to 6K samples, i.e., by 92.9\%. These gains are consistent with the architectural and training-loop changes introduced in Tab.~\ref{tab:policy_comparison} and Sec.~\ref{sec:gb_pl}.
\subsection{Closed-Loop Performance Metrics}
\begin{table}[!t]
    \centering
    \caption{Closed-loop performance and training-sample budget on Monza over \SI{120}{s}. Best, second-best, and third-best entries are highlighted in green, bold, and underline.}
    \label{tab:performance_overview}
    \small
    \setlength{\tabcolsep}{3pt}
    \begin{tabular}{@{}>{\raggedright\arraybackslash}m{2.80cm}@{\hspace{0.02cm}}>{\centering\arraybackslash}m{2.9cm}@{\hspace{0.02cm}}>{\centering\arraybackslash}m{2.9cm}@{}}
    \toprule
    \textbf{Method} & \textbf{\shortstack{Total return $\uparrow$}} & \textbf{\shortstack{Training  samples $\downarrow$}} \\
    \hline
    Human Expert & \shortstack{-993.53} & - \\ \hline
    Reduced \gls{gbpl} \cite{2025GB_PL_DiffNMPC} & \shortstack{-118.62} & \shortstack{\textbf{94K}} \\
    \hline
    \gls{gbpl} (ours) & \shortstack{-81.87 \\ (+31.0\% vs \\red. \glsentryshort{gbpl} (SOTA))} & \cellcolor[HTML]{D9EAD3}\shortstack{\textbf{6K} \\ (-92.9\% vs \\red. \glsentryshort{gbpl} (SOTA))} \\
    \hline
    \gls{rl} PPO (SOTA) & \shortstack{-38.31 \\ (+53.2\% vs \glsentryshort{gbpl})} & \shortstack{1.94M} \\ \hline
    \glsentryshort{sgsca} (ours) & \cellcolor[HTML]{D9EAD3}\shortstack{\textbf{-33.17} \\ (+59.5\% vs \glsentryshort{gbpl})} & \shortstack{\underline{810K} \\ (-58.3\% vs PPO)} \\ \hline
    \glsentryshort{sglos} (ours) & \shortstack{\textbf{-34.69} \\ (+57.6\% vs \glsentryshort{gbpl})} & \shortstack{\underline{810K} \\ (-58.3\% vs PPO)} \\ \hline
    \glsentryshort{sgadv} (ours) & \shortstack{-37.41 \\ (+54.3\% vs \glsentryshort{gbpl})} & \shortstack{972K \\ (-50.0\% vs PPO)} \\ \hline
    \glsentryshort{sgcrt} (ours) & \shortstack{\underline{-35.55} \\ (+56.6\% vs \glsentryshort{gbpl})} & \shortstack{972K \\ (-50.0\% vs PPO)} \\
    \bottomrule
    \end{tabular}
\end{table}
Table~\ref{tab:performance_overview} isolates total return and sample budget for the single-seed AV-24/Monza comparison. All \gls{sgrl} methods outperform our \gls{gbpl} baseline, with \glsentryshort{sgsca} best overall; \glsentryshort{sgsca} and \glsentryshort{sglos} need only 810K samples, 58.3\% fewer than PPO's 1.94M budget.
Lap time is not the primary metric; the optimized objective is Eq.~\eqref{eq:L_perf}, which balances path tracking, velocity tracking, and control smoothness. 

\begin{table}[!t]
    \centering
    \caption{Closed-loop metrics on Monza over \SI{120}{s}: lateral-/velocity deviation $e_{\mathrm{lat}}$/$e_v$, jerk $j$, and steering rate $\dot{\delta}$.}
    \label{tab:closed_loop_metrics_compact}
    \small
    \setlength{\tabcolsep}{2pt}
    \begin{tabular}{@{}>{\raggedright\arraybackslash}m{2.6cm}@{\hspace{0.01cm}}>{\centering\arraybackslash}m{1.15cm}@{\hspace{0.01cm}}>{\centering\arraybackslash}m{1.15cm}@{\hspace{0.01cm}}>{\centering\arraybackslash}m{1.15cm}@{\hspace{0.01cm}}>{\centering\arraybackslash}m{1.15cm}@{\hspace{0.01cm}}>{\centering\arraybackslash}m{1.15cm}@{}}
    \toprule
    \textbf{Method} & \shortstack{$\overline{|e_{\mathrm{lat}}|}$\\ {[m]}} & \shortstack{$\max |e_{\mathrm{lat}}|$\\ {[m]}} & \shortstack{$\overline{|e_v|}$\\ {[m/s]}} & \shortstack{$\max |j|$\\ {[m/s$^3$]}} & \shortstack{$\max |\dot{\delta}|$\\ {[rad/s]}} \\
    \hline
    Human Expert & \shortstack{0.167} & \shortstack{1.50} & \shortstack{0.40} & \cellcolor[HTML]{D9EAD3}\shortstack{\textbf{16.16}} & \shortstack{0.41} \\ \hline
    Reduced \gls{gbpl} \cite{2025GB_PL_DiffNMPC} & \shortstack{0.049} & \shortstack{0.83} & \shortstack{0.25} & \shortstack{127.62} & \shortstack{0.41} \\ \hline
    \glsentryshort{gbpl} (ours) & \shortstack{\underline{0.044}} & \shortstack{\underline{0.66}} & \shortstack{\underline{0.22}} & \shortstack{83.44} & \shortstack{\underline{0.39}} \\ \hline
    \gls{rl} PPO (SOTA) & \shortstack{\textbf{0.035}} & \shortstack{\textbf{0.38}} & \cellcolor[HTML]{D9EAD3}\shortstack{\textbf{0.14}} & \shortstack{\textbf{26.60}} & \shortstack{\textbf{0.35}} \\ \hline
    \glsentryshort{sgsca} (ours) & \cellcolor[HTML]{D9EAD3}\shortstack{\textbf{0.034}} & \cellcolor[HTML]{D9EAD3}\shortstack{\textbf{0.33}} & \shortstack{\textbf{0.15}} & \shortstack{\underline{42.55}} & \cellcolor[HTML]{D9EAD3}\shortstack{\textbf{0.34}} \\
    \bottomrule
    \end{tabular}
\end{table}
Table~\ref{tab:closed_loop_metrics_compact} reports physical closed-loop metrics. We show \glsentryshort{sgsca} as representative because the \gls{sgrl} variants have similar Monza statistics. The fixed-weight expert is smooth in maximum jerk but tracks poorly, while both \gls{gbpl} methods are more aggressive, especially in peak jerk. PPO and \glsentryshort{sgsca} achieve the tightest tracking and lowest maximum steering rates, so \gls{sgrl}'s sample-efficiency gains do not sacrifice closed-loop quality.
\subsection{Generalization in Unseen Environments}
\begin{figure}[!t]
    \centering
    \includegraphics[width=0.85\columnwidth]{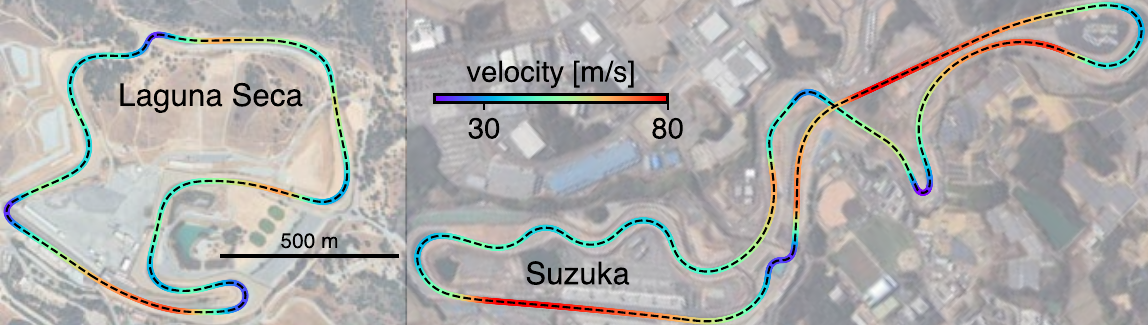}
        \caption{
            Zero-shot evaluation on race tracks unseen during training.
            }
    \label{fig:unseen_tracks}
\end{figure}

Although training uses one circuit per platform, the policies are designed to be \emph{track-agnostic}: they do not receive track coordinates and instead condition weight selection on local state and short-horizon reference information (Sec.~\ref{sec:nn_policy}). We therefore evaluate each learned policy zero-shot on unseen racelines (Fig.~\ref{fig:unseen_tracks}). 
Table~\ref{tab:zeroshot_performance} shows robust transfer and large gains over our white-box \gls{gbpl} baseline on Laguna Seca and Suzuka.
\begin{table}[!t]
    \centering
    \caption{Zero-shot evaluation of trained policies on tracks unseen during training. 
    Each cell reports total return and the relative improvement compared to the \glsentryshort{gbpl} baseline. 
    }
    \label{tab:zeroshot_performance}
    \small
    \setlength{\tabcolsep}{3pt}
    \begin{tabular}{@{}>{\raggedright\arraybackslash}m{2.55cm}@{\hspace{0.02cm}}>{\centering\arraybackslash}m{2.70cm}@{\hspace{0.02cm}}>{\centering\arraybackslash}m{2.70cm}@{}}
    \toprule
    \textbf{Method} & \textbf{\shortstack{Laguna Seca \\Zero-shot}} & \textbf{\shortstack{Suzuka \\Zero-shot}} \\
    \hline
    \glsentryshort{gbpl} baseline & \shortstack{-120.59} & \shortstack{-194.15} \\ \hline
    \gls{rl} PPO baseline & \shortstack{-43.73 \\ (+63.7\% vs \glsentryshort{gbpl})} & \shortstack{-85.81 \\ (+55.8\% vs \glsentryshort{gbpl})} \\ \hline
    \glsentryshort{sgsca} (ours) & \shortstack{\underline{-40.51} \\ (+66.4\% vs \glsentryshort{gbpl})} & \cellcolor[HTML]{D9EAD3}\shortstack{\textbf{-73.52} \\ (+62.1\% vs \glsentryshort{gbpl})} \\ \hline
    \glsentryshort{sglos} (ours) & \shortstack{\textbf{-39.77} \\ (+67.0\% vs \glsentryshort{gbpl})} & \shortstack{\underline{-79.92} \\ (+58.8\% vs \glsentryshort{gbpl})} \\ \hline
    \glsentryshort{sgadv} (ours) & \shortstack{-41.48 \\ (+65.6\% vs \glsentryshort{gbpl})} & \shortstack{-85.25 \\ (+56.1\% vs \glsentryshort{gbpl})} \\ \hline
    \glsentryshort{sgcrt} (ours) & \cellcolor[HTML]{D9EAD3}\shortstack{\textbf{-37.89} \\ (+68.6\% vs \glsentryshort{gbpl})} & \shortstack{\textbf{-78.37} \\ (+59.6\% vs \glsentryshort{gbpl})} \\
    \bottomrule
    \end{tabular}
\end{table}
On Laguna Seca, the PPO baseline already improves on our \gls{gbpl} baseline by 63.7\%, and all four \gls{sgrl} methods improve further to 65.6\%--68.6\%; the strongest transfer result is obtained by \gls{sgcrt} at 68.6\%, followed closely by \gls{sglos} at 67.0\% and \gls{sgsca} at 66.4\%. On Suzuka, PPO improves on our \gls{gbpl} baseline by 55.8\%, whereas \gls{sgsca} reaches 62.1\%, \gls{sgcrt} 59.6\%, \gls{sglos} 58.8\%, and \gls{sgadv} 56.1\%. Notably, these transfer gains use the single-seed AV-24/Monza policies trained with only $8.10\times 10^5$ to $9.72\times 10^5$ interaction samples, i.e., about 50\%--58\% fewer than PPO's $1.94\times 10^6$.

These zero-shot results indicate that the policy learns a transferable map from local driving context to \gls{nmpc} weight rebalancing rather than overfitting to a raceline. \gls{sgrl} maintains, and often improves on, PPO's unseen-track performance.

\subsection{Explaining Cost-Weight Policies via Correlation Structure}
\label{sec:results_explainable_weights}
\begin{figure}[!t]
    \centering
    \includegraphics[width=\columnwidth]{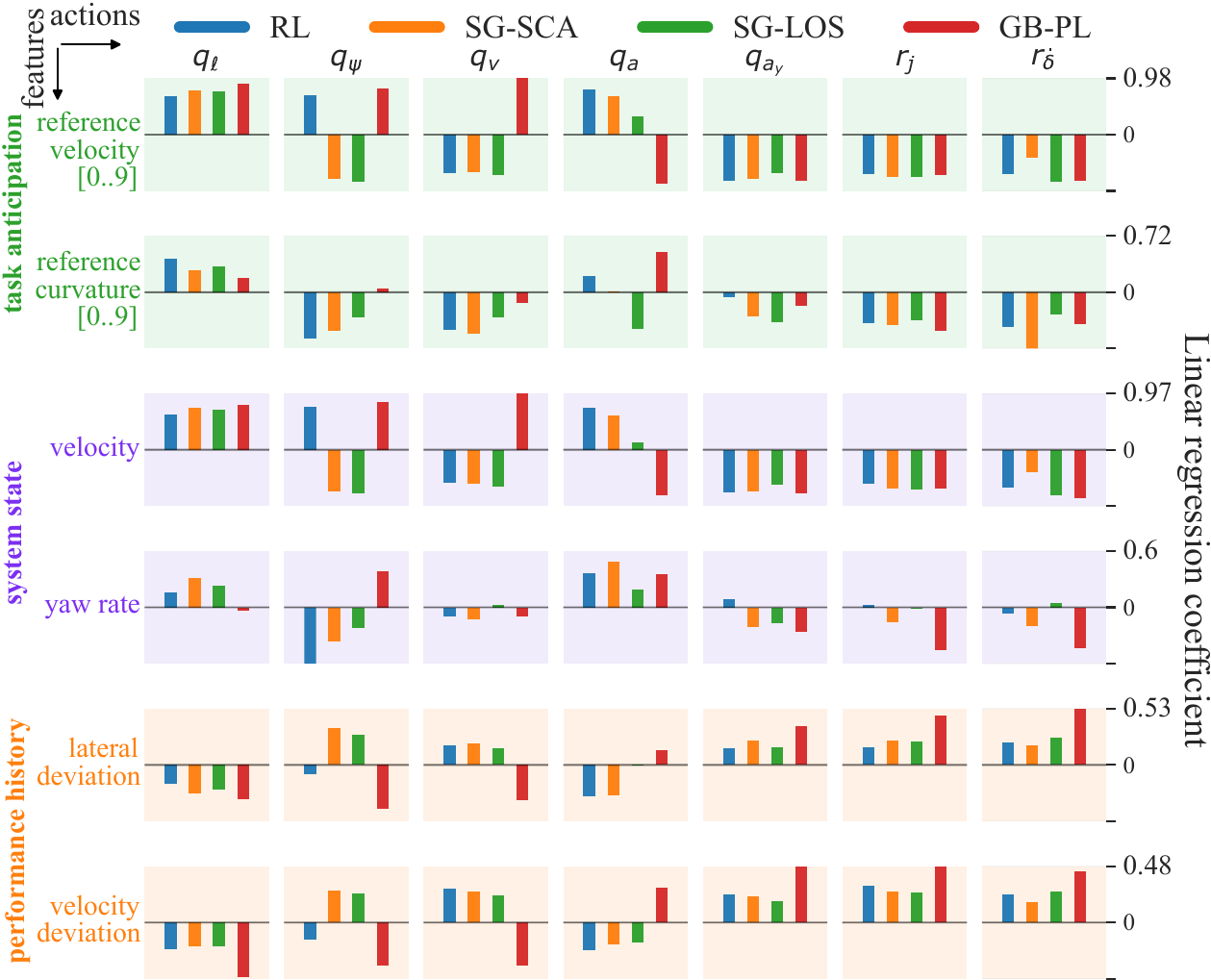}
    \caption{Correlations between learned \gls{nmpc} cost weights and grouped observation features on Monza. Bars compare which signals each policy uses to adapt each weight.}
    \label{fig:correlations_matrix}
\end{figure}
\begin{figure}[!t]
    \centering
    \includegraphics[width=\columnwidth]{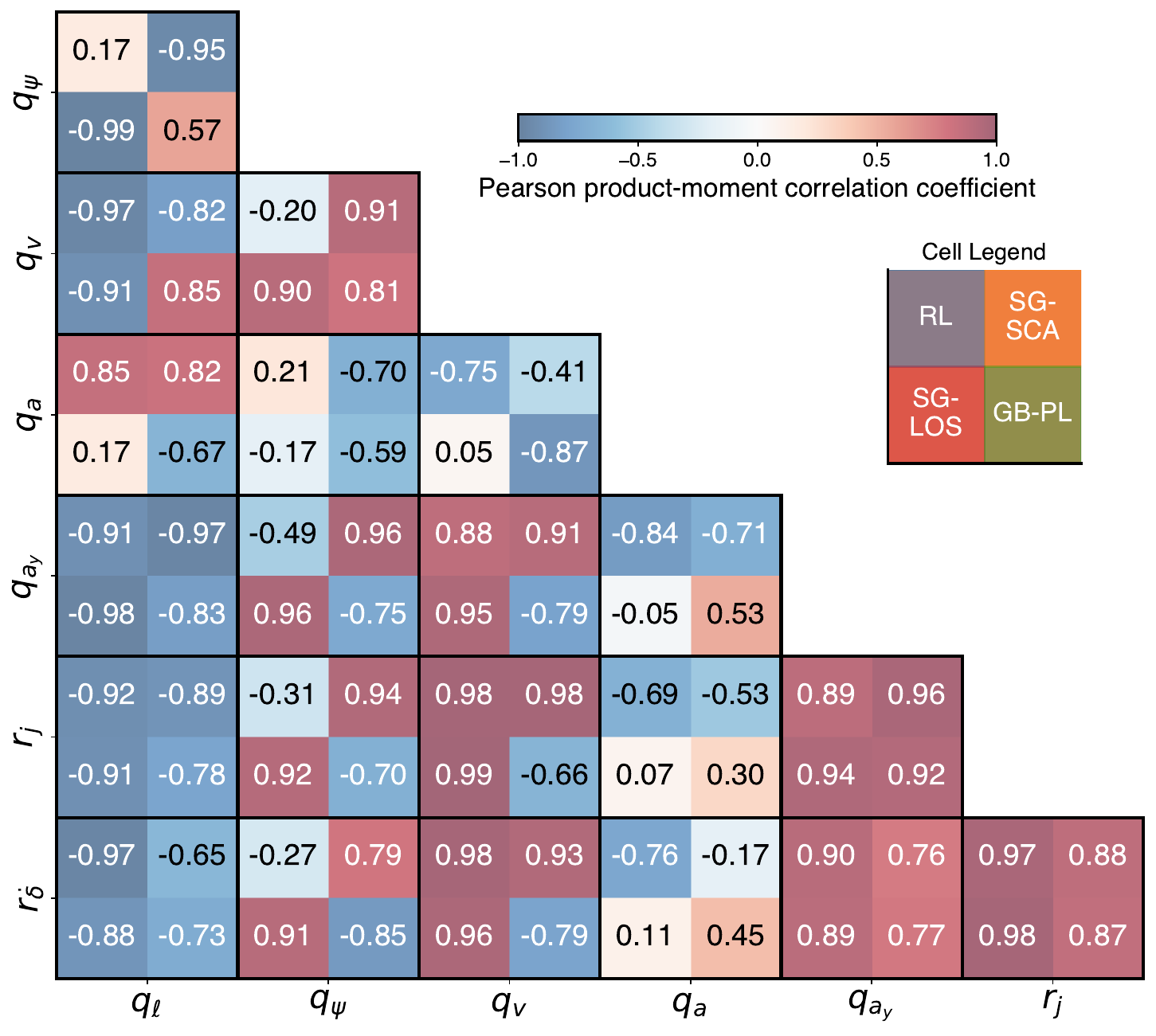}
    \caption{Pairwise correlations between learned \gls{nmpc} cost weights along Monza rollouts. Each sub-cell compares one policy, with color indicating correlation strength.}
    \label{fig:weights_interdependencies}
\end{figure}

Because policies output \emph{cost weights}, not controls, we interpret adaptation via correlations between observation groups and weights, and via inter-weight correlations during rollout. Fig.~\ref{fig:correlations_matrix} reports signed Pearson correlations with six grouped features, while Fig.~\ref{fig:weights_interdependencies} shows weight co-variation. These statistics indicate associations, not causality.
Across methods, the dominant dependencies differ (Figs.~\ref{fig:correlations_matrix}--\ref{fig:weights_interdependencies}). Overall, all policies condition strongly on velocity signals, but solver-gradient guidance changes \emph{which} weights are most strongly tied to anticipation versus instantaneous state. The \gls{gbpl} baseline exhibits a near-direct velocity-aware structure: $q_v$ is almost perfectly aligned with reference velocity ($\rho=0.98$), while $q_a$ anti-correlates with it ($\rho=-0.86$), yielding an optimization-like trade-off also visible as $\rho(q_a,q_v)=-0.87$. The \gls{rl} baseline similarly shows strong velocity conditioning (e.g., $\rho(q_a,\text{ref. velocity})=0.79$ and $\rho(q_{a_y},\text{ref. velocity})=-0.80$), but its interdependencies reveal an almost rank-one ``see-saw'' between aggressiveness/smoothness and lateral correction (e.g., $\rho(r_j,q_v)=0.98$ and $\rho(q_v,q_\ell)=-0.97$).

Solver-gradient guided methods exhibit distinct, method-specific structure. \gls{sgsca} assigns curvature-driven modulation primarily to steering smoothness: $r_{\dot{\delta}}$ correlates more with reference curvature ($\rho=-0.72$) than with reference velocity ($\rho=-0.40$), while lateral tracking weights anti-correlate ($\rho(q_\psi,q_\ell)=-0.95$), indicating a systematic lateral trade-off. \gls{sglos} concentrates its strongest velocity dependence on steering smoothness ($\rho(r_{\dot{\delta}},\text{ref. velocity})=-0.81$) and is the only method where longitudinal aggressiveness is primarily curvature-associated ($\rho(q_a,\text{ref. curvature})=-0.47$), suggesting geometry-aware tightening/relaxation of acceleration shaping. In both guided methods, positive couplings such as $\rho(r_j,q_v)\approx0.98$--$0.99$ indicate coordinated velocity/smoothness scaling, while negative pairs (e.g., $\rho(q_\psi,q_\ell)=-0.99$ for \gls{sglos}) reveal objective rebalancing rather than uniform tightening.

\subsection{Closed-Loop Weight Adaptation Dynamics Along the Track}
\label{sec:results_weight_dynamics}
\begin{figure}[!t]
    \centering
    \includegraphics[width=\columnwidth]{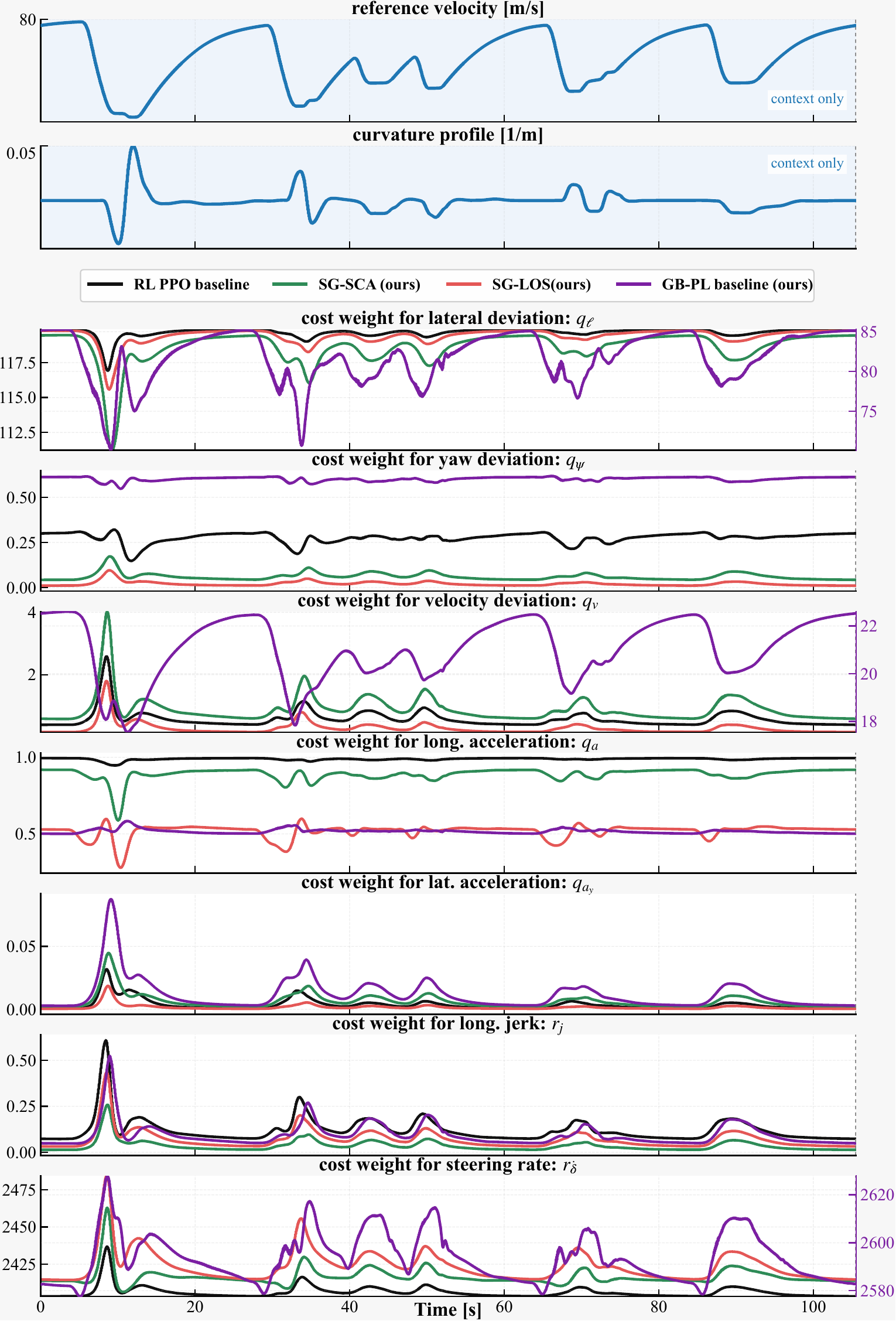}
    \caption{Time evolution of learned cost weights during a Monza lap rollout. Reference velocity and curvature are shown above the seven \gls{nmpc} weights; \glsentryshort{gbpl} uses separate axes where its scale differs strongly.}
    \label{fig:mpc_weights_evolution}
\end{figure}
Correlations identify which signals associate with each weight; Fig.~\ref{fig:mpc_weights_evolution} adds the temporal view of how weights are scheduled along the lap, separating methods more clearly than static summaries.

\textbf{\gls{rl} PPO baseline: coarse but behavior-linked scheduling.}
PPO keeps $q_{\ell}$ and $q_a$ near upper ranges, varies $q_\psi$ moderately, and mainly activates $r_j$, $r_{\dot{\delta}}$, and short $q_v$ bursts around braking/turn-in. Thus, PPO adapts event-wise rather than tracing the reference-velocity template, but through a low-dimensional aggressiveness--smoothness mode.

\textbf{\glsentryshort{sgsca}: sparse, targeted reweighting.}
\glsentryshort{sgsca} preserves PPO's sparse character but shifts interventions: it adds sharper localized $q_\psi$, $q_v$, and $q_{a_y}$ activations near dominant curvature/deceleration zones while keeping jerk and steering-rate regularization restrained. The schedule is selective, reinforcing task-relevant penalties only in specific segments.

\textbf{\glsentryshort{sglos}: coordinated smoothness gating.}
\glsentryshort{sglos} most clearly co-schedules smoothness-related objectives: $q_a$, $r_j$, and $r_{\dot{\delta}}$ rise and fall together, while $q_\psi$ and $q_v$ stay small except for short corrections. This matches training diagnostics: the guidance loss synchronizes updates in performance-critical regions rather than simply increasing PPO aggressiveness.

\textbf{\glsentryshort{gbpl} baseline: velocity-template overfitting.}
\glsentryshort{gbpl} shows the strongest continuous variation in $q_v$ and $r_{\dot{\delta}}$, plus the largest $q_{a_y}$ peaks at high-curvature events. Its $q_v$ trajectory nearly reproduces the reference-velocity profile, suggesting velocity-weight overfitting.
By contrast, \gls{rl} and \gls{sgrl} keep $q_v$ smaller and activate it jointly with lateral/smoothness weights where closed-loop behavior demands it, matching their stronger returns.
\subsection{Ablations: Removing Future and Past Context}
\label{sec:results_ablation_anticipation}
\begin{figure}[!t]
    \centering
    \includegraphics[width=0.99\columnwidth]{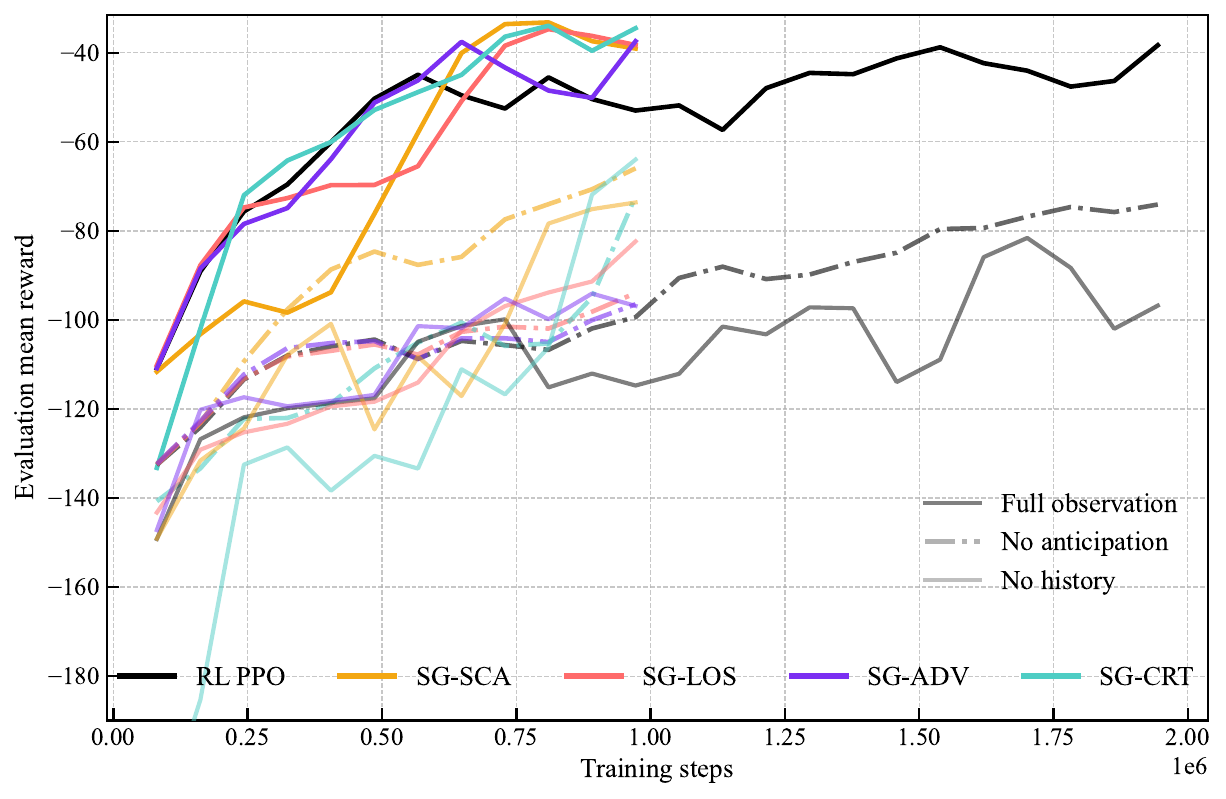}
        \caption{Observation-ablation study for PPO and \gls{sgrl} methods. Curves compare full observations against variants without task anticipation or performance history.}
    \label{fig:sg_sweeps_eval_rewards_ablation_combined}
\end{figure}

Sec.~\ref{sec:nn_policy} assumes that effective adaptation needs two channels: \emph{task anticipation} from reference lookahead and \emph{performance history} for compensating persistent errors or unmodeled dynamics. We test this with two ablations that keep the \gls{nmpc} formulation, constraints, and reference fixed while restricting policy information: \emph{no anticipation} removes horizon-sampled reference velocity/curvature, and \emph{no history} removes recent lateral/velocity deviations. Fig.~\ref{fig:sg_sweeps_eval_rewards_ablation_combined} shows learning curves; Table~\ref{tab:ablation_peak_reward} reports peak-reward degradation.
\subsubsection{Ablation 1 (No anticipation): The Necessity of Preview}
Removing preview information (dashed curves in Fig.~\ref{fig:sg_sweeps_eval_rewards_ablation_combined}) systematically degrades performance: the policy becomes feedback-only and cannot anticipate curvature or deceleration. It therefore modulates weights only \emph{after} entering challenging segments, confirming the anticipation-aware structure in Figs.~\ref{fig:correlations_matrix}--\ref{fig:weights_interdependencies}.
\begin{table}[!t]
    \centering
    \caption{Effect of removing task-anticipation and performance-history features. Each cell reports ablated reward and degradation relative to the full-observation policy.}
    \label{tab:ablation_peak_reward}
    \small
    \setlength{\tabcolsep}{3pt}
    \begin{tabular}{@{}>{\raggedright\arraybackslash}m{2.55cm}@{\hspace{0.02cm}}>{\centering\arraybackslash}m{2.70cm}@{\hspace{0.02cm}}>{\centering\arraybackslash}m{2.70cm}@{}}
    \toprule
    \textbf{Method} & \textbf{\shortstack{No\\ anticipation}} & \textbf{\shortstack{No\\ history}} \\
    \hline
    \glsentryshort{gbpl} baseline & \shortstack{-123.79 \\ (-51.2\% vs full)} & \shortstack{-94.08 \\ (-14.9\% vs full)} \\ \hline
    \gls{rl} PPO baseline & \shortstack{\underline{-74.07} \\ (-93.4\% vs full)} & \shortstack{\underline{-81.58} \\ (-113.0\% vs full)} \\ \hline
    \glsentryshort{sgsca} (ours) & \cellcolor[HTML]{D9EAD3}\shortstack{\textbf{-65.95} \\ (-98.8\% vs full)} & \shortstack{\textbf{-73.61} \\ (-121.9\% vs full)} \\ \hline
    \glsentryshort{sglos} (ours) & \shortstack{-93.96 \\ (-170.9\% vs full)} & \shortstack{-82.36 \\ (-137.4\% vs full)} \\ \hline
    \glsentryshort{sgadv} (ours) & \shortstack{-96.49 \\ (-157.9\% vs full)} & \shortstack{-94.08 \\ (-151.5\% vs full)} \\ \hline
    \glsentryshort{sgcrt} (ours) & \shortstack{\textbf{-72.20} \\ (-103.1\% vs full)} & \cellcolor[HTML]{D9EAD3}\shortstack{\textbf{-63.98} \\ (-80.0\% vs full)} \\
    \bottomrule
    \end{tabular}
\end{table}
Among guided methods, \emph{\glsentryshort{sgsca}} is most robust to losing anticipation, while \emph{\glsentryshort{sgcrt}} still outperforms \gls{rl}, suggesting gradient-informed value estimation remains useful when feedforward context is reduced. 

\subsubsection{Ablation 2 (No history): The Role of Temporal Context}
Eliminating performance-history features (solid curves) also degrades reward, but more heterogeneously than removing anticipation. PPO and \glsentryshort{sgsca} suffer larger losses without history than without preview, while \glsentryshort{sglos}, \glsentryshort{sgadv}, and \glsentryshort{sgcrt} retain more of their full-observation performance than in the no-anticipation case. With only a memoryless state snapshot, the policy cannot distinguish transient execution errors from persistent performance deviations. As a result, it cannot reliably modulate the persistence or magnitude of weight adjustments, which impairs closed-loop consistency over extended operation. This aligns with Sec.~\ref{sec:results_explainable_weights}, where deviation-history terms strongly correlate with regulatory weights.

Our \gls{gbpl} baseline degrades only modestly, consistent with its myopic update: it descends a local sensitivity gradient $\nabla_{\boldsymbol{\theta}}\mathcal{L}$ on the current trajectory and depends less on history. This robustness costs optimality, since \gls{gbpl} remains below the best SG-guided methods in absolute reward (Table~\ref{tab:ablation_peak_reward}).


\textbf{Advantage of solver gradients under information scarcity.}
As Table~\ref{tab:ablation_peak_reward} shows, the strongest SG-guided methods continue to outperform the pure \gls{ppo} baseline in peak reward even after we remove temporal context.
\subsection{Training-Dynamics Analysis}
\paragraph{Core \gls{ppo} training dynamics in the windows where \gls{sgrl} leads.}
To explain why certain \gls{sgrl} methods accelerate \gls{ppo}, we inspect \gls{ppo} signals in an early \emph{acceleration window}, where a method first pulls ahead, and a later \emph{peak window}, where it reaches best performance. Table~\ref{tab:ppo_window_dynamics} compares the winning SG method with \gls{ppo} in these windows.
\begin{table}[!t]
\centering
\caption{\textbf{Core \gls{ppo} signals in performance-relevant windows.} Entries are window means for the winning \gls{sgrl} method and \gls{ppo} baseline, reported as SG/\gls{ppo}. KL and clip fraction indicate trust-region activity; value, policy-gradient, and entropy losses summarize critic fit, actor update magnitude, and exploration; weight/bias norms refer to the actor action head.}
\label{tab:ppo_window_dynamics}
\resizebox{\columnwidth}{!}{
\scriptsize
{\setlength{\tabcolsep}{1pt}
\renewcommand{\arraystretch}{1.05}
\begin{tabular}{l | c c | c c}
\hline
\multirow{3}{*}{Signal} & \multicolumn{2}{c|}{AV-24 Monza} & \multicolumn{2}{c}{EAV-24 Yas Marina} \\
\cline{2-5}
 & \shortstack{Acceleration window\\ winner: \glsentryshort{sglos}} & \shortstack{Peak window\\ winner: \glsentryshort{sgsca}} & \shortstack{Acceleration window\\ winner: \glsentryshort{sgsca}} & \shortstack{Peak window\\ winner: \glsentryshort{sglos}} \\
\hline
Eval return & $-67.3 / -81.6$ & $-44.0 / -68.4$ & $-93.6 / -111$ & $-86.4 / -103$ \\
KL & $0.0935/0.161$ & $0.0337/0.0721$ & $0.107/0.164$ & $0.0567/0.0657$ \\
Clip frac. & $0.309/0.366$ & $0.225/0.314$ & $0.337/0.377$ & $0.254/0.284$ \\
Value loss & $296/135$ & $1.51\mathrm{e}3/502$ & $0.750/0.774$ & $2.55/6.33$ \\
$|$PG loss$|$ & $0.0328/0.0629$ & $9.48\mathrm{e}{-3}/0.0307$ & $0.0466/0.0641$ & $0.0165/0.0247$ \\
$|$Entropy loss$|$ & $8.31/9.26$ & $6.99/8.37$ & $8.79/9.16$ & $6.63/7.57$ \\
Weight norm & $2.67/2.65$ & $2.69/2.66$ & $2.67/2.66$ & $2.73/2.69$ \\
Bias norm & $7.23/1.99\mathrm{e}{-2}$ & $7.23/2.44\mathrm{e}{-2}$ & $1.63/1.51$ & $1.47/1.51$ \\
\hline
\end{tabular}
}
}
\end{table}
Table~\ref{tab:ppo_window_dynamics} shows that winning SG methods beat \gls{ppo} through \emph{more structured}, not indiscriminately larger, updates: they regularize \gls{ppo} toward better-timed and lower-variance updates. 

In AV-24/Monza, \gls{sglos} wins the early phase, improving the windowed return from $-81.6$ to $-67.3$ compared to \gls{ppo}, while reducing KL from $0.161$ to $0.0935$ and clip fraction from $0.366$ to $0.309$. The peak window is then won by \gls{sgsca}, which lifts the mean return from $-68.4$ to $-44.0$ with roughly half the \gls{ppo} KL ($0.0337$ vs. $0.0721$), a lower clip fraction, and a much smaller $|$PG loss$|$. On this platform, the action-head weight norm remains close to \gls{ppo}, while the bias norm separates more strongly for the winning SG methods.

In EAV-24/Yas Marina, the lead changes by phase: \gls{sgsca} wins the acceleration window and \gls{sglos} wins the peak window. During training, \gls{sgsca} improves mean return from $-111$ to $-93.6$ while lowering KL, clip fraction, value loss, $|$PG loss$|$, and $|$entropy loss$|$. Near peak performance, \gls{sglos} improves from $-103$ to $-86.4$ and again lowers KL, clip fraction, value loss, $|$PG loss$|$, and $|$entropy loss$|$. 

\paragraph{Method-specific guidance in performance-relevant windows.}
Table~\ref{tab:sg_method_window_diagnostics} resolves the window winners into their active guidance channels. 
\begin{table}[!t]
\centering
\caption{\textbf{Method-specific \gls{sgrl} signals in performance-relevant windows.} Entries are window means in the same windows as Table~\ref{tab:ppo_window_dynamics}. For \glsentryshort{sgsca}, $\bar c/\bar s/\bar f_{\mathrm{clamp}}$ denote alignment, projected scale, and clamp-hit fraction; for \glsentryshort{sglos}, $\bar\ell_{\mathrm{guide}}$ is the guide loss; for \glsentryshort{sgadv}, $\bar\rho_{\mathrm{AS}}/\rho_{\mathrm{AS}}^{90}$ quantify advantage correction; for \glsentryshort{sgcrt}, $\bar\rho_V$ is the critic correction ratio.}
\label{tab:sg_method_window_diagnostics}
\resizebox{\columnwidth}{!}{
\scriptsize
{\setlength{\tabcolsep}{2pt}
\renewcommand{\arraystretch}{1.0}
\begin{tabular}{l | c c | c c}
\hline
Method / signal & \multicolumn{2}{c|}{AV-24 Monza} & \multicolumn{2}{c}{EAV-24 Yas Marina} \\
\cline{2-5}
 & Accel. & Peak & Accel. & Peak \\
\hline
\glsentryshort{sgsca}: $\bar c$ & $-1.42\mathrm{e}{-3}$ & $-0.0140$ & $8.02\mathrm{e}{-3}$ & $0.0215$ \\
\glsentryshort{sgsca}: $\bar s$ & $0.986$ & $0.944$ & $1.05$ & $1.05$ \\
\glsentryshort{sgsca}: $\bar f_{\mathrm{clamp}}$ & $0.527$ & $0.168$ & $0.541$ & $0.452$ \\
\glsentryshort{sglos}: $\bar\ell_{\mathrm{guide}}$ & $2.69\mathrm{e}{-3}$ & $1.98\mathrm{e}{-3}$ & $3.61\mathrm{e}{-3}$ & $2.89\mathrm{e}{-3}$ \\
\glsentryshort{sgadv}: $\bar\rho_{\mathrm{AS}}$ & $0.336$ & $0.447$ & $0.0116$ & $0.0155$ \\
\glsentryshort{sgadv}: $\rho_{\mathrm{AS}}^{90}$ & $0.356$ & $0.499$ & $0.0118$ & $0.0159$ \\
\glsentryshort{sgcrt}: $\bar\rho_V$ & $1.99$ & $1.34$ & $0.0451$ & $0.0478$ \\
\hline
\end{tabular}
}
}
\end{table}
On AV-24/Monza, \glsentryshort{sglos} keeps nonzero acceleration-window guidance ($\bar\ell_{\mathrm{guide}}=2.69\times10^{-3}$), so early gains reflect solver-biased shaping while PPO is noisy. For \glsentryshort{sgsca}, peak-window $\bar c$ becomes more negative, $\bar s<1$, and $\bar f_{\mathrm{clamp}}$ drops from $0.527$ to $0.168$, indicating selective braking of PPO steps that drift from solver-informed directions rather than persistent acceleration or repeated clamp saturation.

On EAV-24/Yas Marina, \glsentryshort{sgsca}'s sign pattern flips: $\bar c$ stays positive and grows from $8.02\times10^{-3}$ to $0.0215$, with $\bar s\approx1.05$, so it nudges PPO along solver-aligned directions and wins the acceleration window. \glsentryshort{sglos} remains strong at peak because its guidance term stays large ($3.61\times10^{-3}$ early, $2.89\times10^{-3}$ at peak), continuing to shape the actor objective. \glsentryshort{sgadv} and \glsentryshort{sgcrt} mainly support platform-specific gains: large $\bar\rho_{\mathrm{AS}}$ or $\bar\rho_V$ indicate strong advantage/critic reshaping, which is pronounced on Monza but weak on Yas Marina.

\section{CONCLUSIONS}

We introduced \textbf{Solver-Gradient Guided Reinforcement Learning (\gls{sgrl})}, a framework for online \gls{mpc} cost-weight adaptation that combines environment-based policy-gradient learning with model-based solver-gradient guidance while preserving the realized closed-loop return as the optimization objective. Our \gls{sgrl} uses solver sensitivities as auxiliary guidance signals for \gls{rl} to improve policy learning.

We instantiated this framework through four methods---\gls{sgsca}, \gls{sglos}, \gls{sgadv}, and \gls{sgcrt}---that incorporate solver guidance into different components of PPO. Across two full-scale autonomous racing platforms under intentional model mismatch, all four methods consistently improved sample efficiency and closed-loop performance over PPO, matched PPO's best return with up to 70.6\,\% fewer training samples, outperformed \gls{gbpl} baselines by at least 54\,\% in return, and generalized zero-shot to unseen race tracks. Among the proposed methods, \gls{sgsca} and \gls{sglos} achieved the strongest overall performance.

Overall, our results indicate that environment-based policy gradients and model-based solver guidance are complementary optimization signals for learning online \gls{mpc} weight-adaptation policies. More broadly, they demonstrate the potential of combining environment interaction with differentiable optimal-control structure for learning in constrained optimal-control problems.

\paragraph*{Limitations}
The effectiveness of \gls{sgrl} depends on the fidelity of the predictive \gls{mpc} model. Under severe model mismatch, solver guidance may become less informative or even misleading for optimizing the true closed-loop objective. Although our experiments include intentional model mismatch and demonstrate robust improvements, they do not systematically study different mismatch types or magnitudes. In addition, each \gls{sgrl} variant introduces method-specific hyperparameters, and this work considers only on-policy PPO.

\paragraph*{Future work}
Future work includes validating the learned weight-adaptation policies on a physical vehicle, extending solver-gradient guidance to off-policy algorithms such as SAC and TD3, and systematically studying robustness under controlled model-mismatch regimes. Another promising direction is to generalize \gls{sgrl} beyond \gls{nmpc} cost-weight adaptation to other differentiable parametric controllers. Finally, developing a stronger theoretical understanding of how solver guidance interacts with stochastic policy-gradient optimization remains an important open research direction.









\bibliographystyle{IEEEtran}
\bibliography{references}

\end{document}